%% file: main.tex
\documentclass[letterpaper]{article} 
\usepackage{aaai24}  
\usepackage{times}  
\usepackage{helvet}  
\usepackage{courier}  
\usepackage[hyphens]{url}  
\usepackage{graphicx} 

\usepackage{paralist}
\usepackage[utf8]{inputenc} 
\usepackage[T1]{fontenc}    
\usepackage{url}            
\usepackage{booktabs}       
\usepackage{amsfonts}       
\usepackage{nicefrac}       
\usepackage{microtype}      
\usepackage{paralist} 
\usepackage[english]{babel}

\usepackage{csquotes}

\usepackage{amsmath}
\usepackage{amssymb}
\usepackage{placeins}
\usepackage{subcaption} 
\usepackage[dvipsnames]{xcolor}
\usepackage{tikz,tkz-euclide,pgfplots}
\pgfplotsset{compat=1.15}
\newtheorem{theorem}{Theorem}

\newcommand{\eg}{e\/.\/g\/.,\/~}

\newcommand*{\parencite}{\citep}
\usepackage{natbib}  
\usepackage{caption} 
\usepackage{algorithm}
\usepackage{algorithmic}
\usepackage[symbol]{footmisc}

\usepackage{newfloat}
\usepackage{listings}
\DeclareCaptionStyle{ruled}{labelfont=normalfont,labelsep=colon,strut=off} 
\floatstyle{ruled}
\newfloat{listing}{tb}{lst}{}
\floatname{listing}{Listing}
\title{ Exact Inference for Continuous-Time Gaussian Process Dynamics}
\author {
	Katharina Ensinger \textsuperscript{\rm{1,2}} \thanks{Authors contributed equally to this work},
	Nicholas Tagliapietra \textsuperscript{\rm 1,$\dagger$,*},
	Sebastian Ziesche \textsuperscript{\rm 1},
	Sebastian Trimpe \textsuperscript{\rm 2}
}
\affiliations {
	\textsuperscript{\rm 1} Bosch Center for Artificial Intelligence, Renningen, Germany \\
	\textsuperscript{\rm 2} Institute for Data Science in Mechanical Engineering, RWTH Aachen University \\ 
	katharina.ensinger@bosch.com
}

\usepackage{bibentry}

\begin{document}

\maketitle
\def\thefootnote{$*$}\footnotetext{Work was conducted at Bosch Center for Artificial Intelligence during master's studies at University of Padova.}
\begin{abstract}
	Many physical systems can be described as a continuous-time dynamical system.
	In practice, the true system is often unknown and has to be learned from measurement data. 
	Since data is typically collected in discrete time, e.g. by sensors, most methods in Gaussian process (GP) dynamics model learning are trained on one-step ahead predictions. 
	While this scheme is mathematically tempting, it can become problematic in several scenarios, e.g. if measurements are provided at irregularly-sampled time steps or physical system properties have to be conserved.
	Thus, we aim for a GP model of the true continuous-time dynamics. 
    We tackle this task by leveraging higher-order numerical integrators. 
    These integrators provide the necessary tools to discretize dynamical systems with arbitrary accuracy.
	However, most higher-order integrators require dynamics evaluations at intermediate time steps, making exact GP inference intractable.
	In previous work, this problem is often addressed by approximate inference techniques.  
	However, exact GP inference is preferable in many scenarios, e.g. due to its mathematical guarantees.
	In order to enable direct inference, we propose to leverage multistep and Taylor integrators. 
	We demonstrate how exact inference schemes can be derived for these types of integrators.
	Further, we derive tailored sampling schemes that allow one to draw consistent dynamics functions from the posterior.
	The learned model can thus be integrated with arbitrary integrators, just like a standard dynamical system. 
	We show empirically and theoretically that our approach yields an accurate representation of the continuous-time system.
\end{abstract}

\input{intro}
\input{background}

\input{method}
\input{related_work}

\input{exp}

\input{conclusion}

\section*{Acknowledgements}
We thank Friedrich Solowjow for valuable discussions. 
\bibliography{bib}

\end{document}


\maketitle

\input{document}

\input{proofs}
\input{add_results}

\input{exp_setup}

\newpage
\bibliography{bib}

 

%% file: intro.tex
\section{Introduction}\label{sec:intro}
Many systems can be described by a continuous-time ordinary differential equation (ODE)
\begin{equation}\label{eq:dyn}
\dot{x}(t)=f(x(t)) \textrm{ with } f:\mathbb{R}^d \rightarrow \mathbb{R}^d.
\end{equation}
Dynamics model learning deals with the problem of estimating the dynamics function $f$ from data.
Usually, it is not possible to measure the states $(x_n)_{n=1 \dots N}$ in continuous time, but noisy measurements $(\hat x_n)_{n=1 \dots N}$ at (potentially irregularly-sampled) discrete time points $t_{n}$ can be obtained by sensors, where
\begin{equation} \label{eq:noisy-data}
\hat{x}_{n,u}=x_{n,u}+\nu_{n,u},
\end{equation}
$\nu_{n,u}\sim \mathcal{N}(0,\sigma_{u}^2)$, $n=1 \dots N$ and $u=1 \dots d$. 
Gaussian processes (GPs) are a powerful probabilistic framework and can be interpreted as a distribution over functions. 
Hence, training a GP is a popular approach in dynamics model learning, providing uncertainty estimates for the model.  
Typically, a GP model $\tilde f$ is trained on one-step ahead predictions \cite{10.5555/3104482.3104541, doerr2018probabilistic}
\begin{equation}\label{eq:oneStep}
x_{n+1}=\tilde f(x_n).
\end{equation}
Mathematically, this structure is tempting since it can be addressed via standard GP regression, where input-output pairs are given as neighboring points in the trajectory. 
Thus, the GP can be directly conditioned on the data. 
However, the approach can lead problems in various scenarios.
Especially, it is not suitable for irregularly-sampled or missing training data or if predictions at intermediate time steps are required. 
Further, physical structure of the true system, such as energy or volume is typically not preserved \citep{https://doi.org/10.48550/arxiv.2102.01606}.

We address this problem by learning GP dynamics that represent $f$ more accurately.
To this end, we leverage numerical integrators, in particular higher-order multistep and Taylor integrators.
Numerical integrators provide the necessary tools to approximate the solution of system \eqref{eq:dyn} with arbitrary accuracy for known dynamics $f$. 
However, we can leverage the approximation qualities of numerical integrators in a learning-based scenario, where the true dynamics are unknown. 
In particular, applying a higher-order integrator to the unknown system Eq. \eqref{eq:dyn} yields a more accurate approximation of the continuous-time dynamics $f$. 
From a numerical perspective, one-step ahead predictions \eqref{eq:oneStep} correspond to the explicit Euler method.
This follows by identifying $\tilde f(x_n)$ with $x_n+(x_{n+1}-x_n)f(x)$.
Being a method of order one, this indicates that the learned dynamics typically do not provide an accurate representation of the continuous-time system.

Generalizing to higher-order integrators raises technical difficulties. 
In particular, most integrators require dynamics evaluations at intermediate time steps \citep{hairerVol1} making standard GP inference intractable.
In previous works, this problem is often addressed by approximating the GP posterior with variational inference \citep{pmlr-v180-hegde22a,  https://doi.org/10.48550/arxiv.2102.01606}.
However, variational inference has some downsides. 
Especially mathematical guarantees of the GP, such as error bounds, get lost. 
The beneficial structure of (varying step-size) multistep and Taylor integrators enables us to derive exact inference schemes.
This is due to the fact that the dynamics are evaluated only at past and current points in the trajectory for these integrators. 
We are thus able to maintain the mathematical properties of standard GP dynamics model learning (cf. Eq. \eqref{eq:oneStep}) while learning an accurate ODE model. 

On a technical level, we derive a flexible framework that automatically computes tailored GP kernels from a given time discretization and integration scheme. 
These kernels are then used for inference.
Further, we derive corresponding decoupled sampling (DS) schemes based on \citep{wilson2020efficiently}.
This technique enables sampling full dynamics functions from the GP posterior.
In contrast to standard GP sampling, is is not required to condition dynamics evaluations on previous ones.
Thus, a trajectory sample is simply obtained by sampling a dynamics function from the posterior and integrating it numerically.
We show theoretically and empirically that we are able to learn good ODE representations by training with integrators of sufficient order.
In summary, our contributions are
\begin{compactitem}
	\item a method that allows for learning GP-based ODE dynamics via standard GP inference enabling the computation of error bounds between true and learned dynamics;
	\item a framework that allows for computing the corresponding kernels for multistep and Taylor integrators of arbitrary order and potentially irregularly-sampled time points; and
	\item a decoupled sampling scheme that allows one to sample consistent dynamics from the GP posterior. 
\end{compactitem} 
 
\begin{figure*}
\centering
\includegraphics[width=0.92\textwidth]{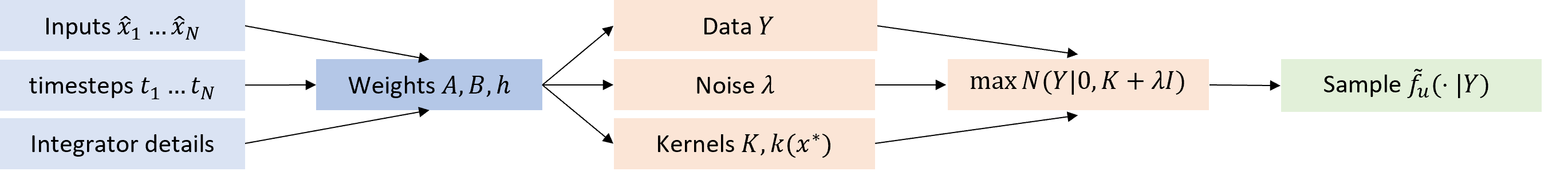}
\caption{Overview over training and predicting for dimension $u$. Data $\hat x_n$, time steps $t_n$ and integrator details are used to generate the integrator coefficients (blue). This allows one to compute all necessary components for training, including transformed observations $Y$, corresponding noise $\lambda$ and kernels (orange). After training, we obtain $\tilde f_u$ via DS (green).}
\label{fig:scheme}
\end{figure*}

%% file: background.tex
\section{Technical Background}
\label{section:background}
We provide a summary of the necessary mathematical tools.
\subsection{Gaussian Process Regression}
\label{section:varGP}
A GP is a random function $g:\mathbb{R}^d \rightarrow \mathbb{R}$ \parencite{10.5555/1162254}. 
Similar to a normal distribution, a GP is determined by its mean function $m(x):= \mathbb{E}[g(x)]$ and covariance function $k(x, y):=\text{cov}(g(x), g(y))$. Here, we assume $m(x)=0$.
Conditioning a GP on observations $Y=(Y_n)_{n=1 \dots N}$ where $Y_n \in \mathbb{R}$, inputs $X = (X_n)_{n=1 \dots N}$ where $X_n \in \mathbb{R}^{d}$ and assuming observation noise distributed with $\mathcal{N}(0,\lambda)$, yields a normal distributed predictive posterior distribution.  
For $M$ test points $(x^{\star}_m)_{m=1 \dots M}$, we obtain  $K:=(k(X_n, X_m))_{n,m =1 \dots N}$, $k(x^\star, x^\star) := (k(x^\star_n, x^\star_m))_{n,m = 1 \dots M}$ and $k(x^{\star}):=(k(x^{\star}_m,X_n))_{m=1 \dots M,n=1 \dots N}$.
Then, it holds that $f(x^{\star}) \sim \mathcal{N}(\mu(x^{\star}),\Sigma(x^{\star}))$ with 
\begin{equation} \label{eq:sparse-moments}
\begin{aligned}
\mu(x^*) &= k(x^{\star})^T(K+\lambda I)^{-1}Y \\ \Sigma(x^{\star})&=k(x^{\star},x^{\star})- k(x^{\star})^T(K+\lambda I)^{-1}k(x^{\star}).
\end{aligned}
\end{equation}
\paragraph{Training: }
During training, we parametrize $k$ with $\theta$ and write $k_{\theta}(x,y)$.  
The trainable parameters $\theta$ and $\lambda$ are adapted by maximizing the log probability of the observations
\begin{equation}\label{eq:loglikeli}
p(Y)=\mathcal{N}(Y|0,K_{\theta}+\lambda I). 
\end{equation}
Remark: In this work, we learn a dynamics function with $d$-dimensional outputs.
This problem is addressed by learning $d$ separate GPs with individual trainable parameters.
\paragraph{Decoupled sampling (DS): }
Standard GP conditioning allows one to evaluate the posterior at a finite subset of points. 
However, trajectory sampling from a learned vector field is an iterative process, where subsequent trajectory points depend on previous ones.
This requires conditioning on previous samples \citep{20.500.11850/454929}, which is intractable for long trajectories and complex integrators. 
To address this problem, decoupled sampling (DS) offers a technique for efficiently drawing consistent functions from the GP posterior \parencite{wilson2020efficiently}.
On a technical level, it decomposes the posterior via Matheron's rule \citep{howarth1979}.
A sample from the posterior is then obtained by combining a sample from the prior with a deterministic update.
The prior can be approximated via a finite-dimensional representation with random basis functions and weights. 
In summary it holds that 
\begin{equation}\label{eq:matheron}
\begin{aligned}
g(x^{\star}|Y)&\approx \sum_{i=1}^{S} w_i \phi_i(x^{\star})+k(x^{\star})(K+\lambda I)^{-1}\\
&(Y-\sum_{i=1}^S w_i\Phi_i-\epsilon)
\end{aligned}
\end{equation}
for the posterior conditioned on observations. Here, the stationary GP prior is represented via $S$ Fourier bases $\phi_i$, $w_i \sim \mathcal{N}(0,1)$ and $\Phi_i=(\Phi_{i,n})_{n=1 \dots N} \in \mathbb{R}^N$ with $\Phi_{i,n}=\phi_i(X_n)$ \parencite{Rahimi}. 
Further, $\epsilon \sim \mathcal{N}(0,\lambda I)$.
\subsection{Numerical Integration}
Numerical integrators for an ODE \eqref{eq:dyn} compute an approximation $\bar{x}_n$ of the solution $x(t_n)$ at discrete time steps $t_n$ \citep{hairerVol1}.
Among other properties, they are determined by their order (of consistency) $P$.
Mathematically, the order corresponds to the truncation index of the Taylor series up to which the correct solution and the approximate solution coincide. 
Therefore, a higher order typically leads to a more accurate approximation $\bar x_n$. 
Here, we use the notation $\bar x_n$ to indicate the subtle difference to ground truth states $x_n$.
\paragraph{Varying step-size multistep integrators: }
\label{section:RK}
Multistep integrators approximate the solution $x(t_{n+M})$ by taking the last $M-1$ points $\bar x_{n}, \dots \bar x_{n+M-1}$ into account. 
It holds that  
\begin{equation}\label{eq:var_step}
\sum_{j=0}^M a_{jn}\bar{x}_{n+j}=\sum_{j=0}^M b_{jn}f(\bar{x}_{n+j}),
\end{equation} 
where the coefficients $A=(a_{jn})_{j=0,\dots, M}^{n=0, \dots, N}$ and $B=(b_{jn})_{j=0, \dots, M}^{n=0,\dots,N}$ depend on the step sizes $h_n=t_{n+1}-t_n$ and $(\bar{x}_{0},\dots, \bar{x}_M) = (x(t_0), \dots x(t_M))$.
In case $b_{M n} \neq 0$, the system is implicit, and thus a minimization problem has to be solved for $\bar x_{n+M}$.  
The parameters $a_{jn}, b_{jn}  \in \mathbb{R}$ determine the properties of the method, \eg the order $P$. 
For constant step sizes, the coefficients $a_{jn}$ and $b_{jn}$ reduce to $a_j$ and $b_j$. 

\paragraph{Taylor integrators: }
Taylor integrators are based on a Taylor expansion of the solution $x(t+h)$ around $x(t)$.
Truncating the expansion at index $P$ yields the Taylor integrator of order $P$. This results in 
\begin{equation}\label{eq:Taylor}
\bar x_{n+1}=\bar x_n +\sum_{l=1}^{P} \frac{h_n^l}{l!}f^l(\bar x_n), 
\end{equation}
where $f^l:\mathbb R^d \rightarrow \mathbb R^d$, $f^1=f, f^2=f^{\prime}f, \dots$, $f^{k+1} =\frac{d}{dx}f^k f$ and $\bar x_0=x(t_0)$ \citep{taylor}. 
The Taylor integrator of order 1 corresponds to the explicit Euler method. 
In practice, $f^l$ is often too complex to compute since it requires the computation of higher-order derivatives of the dynamics $f$. 
However, in a learning-based scenario, the approach is beneficial as we shall see.

%% file: method.tex
\section{Method} \label{method}
Next, we develop the technical details of merging GPs with multistep and Taylor integrators.

\paragraph{Main idea and setup: }
Our goal is to learn an accurate representation of the true \emph{continuous-time} dynamics $f$ from discrete-time trajectory data $\hat x_n$ via exact inference. 
To this end, we apply appropriate numerical integrators, such as multistep and Taylor integrators to the unknown dynamical system (cf. Eq. \eqref{eq:dyn}) and train a GP model for the dynamics.
In the multistep case, we aim for GP dynamics $\tilde f$ that fulfill $\sum_{j=0}^M a_{jn}x_{n+j}=\sum_{j=0}^M b_{jn}\tilde f({x}_{n+j})$.
In the Taylor case, we aim for GP dynamics $\tilde f^l$ that fulfill  $x_{n+1}=x_n+\sum_{l=1}^{P} \frac{h^l}{l!}\tilde f^l(x_n)$.
Intuitively, the higher the order, the more accurate the approximation.

\paragraph{Common setup: }
We model each dimension of the dynamics via separate GPs $\tilde f_u$ with kernel function $k$ in the multistep case for $u=1 \dots d$ and $\tilde f_u^l$ with kernel function $k_l$ in the Taylor case for $u=1 \dots d$ and $l=1 \dots =P$.
For brevity, we omit the index $u$ for the kernels.
The full dynamics are obtained via $\tilde f = (\tilde f_1 \dots \tilde f_d)^T$, respectively $\tilde f^l = (\tilde f_1^l \dots \tilde f_d^l)^T$.
An overview of the method is given in Fig. \ref{fig:scheme}, which we will detail in the following.
We do not model input noise as it is typical for GP dynamics model learning \citep{10.5555/3104482.3104541}.
Therefore, we treat the noisy data $\hat x_n$ as inputs.
However, input noise could be incorporated as proposed by \citet{NIPS2011_a8e864d0}.

\subsection{Data Processing and Kernels} \label{sec:multistep} 
In contrast to the standard scenario, the training data are not given as input-output pairs.
Instead, linear combinations of dynamics evaluations correspond to linear combinations of trajectory points (cf. Eq. \eqref{eq:var_step} and \eqref{eq:Taylor}).
Therefore, the GP observations $Y$ are a linear combination of trajectory points.
Here, we approximate the corresponding time-varying noise $\epsilon \sim \mathcal{N}(0,\lambda I)$ with a diagonal covariance matrix for computational efficiency and stability.
However, our framework provides the option to consider correlated noise between different observations.  
See the appendix for details.   
Similar to the standard regression setting, we compute the covariance matrices $K$ between training inputs and $k(x^{\star})$ between training and test inputs.  
Since all calculations are done dimension wise, we obtain $d$ different representations of $K$ and $k(x^{\star})$.

\paragraph{Multistep integrators: }
We obtain observations $Y=(Y_n)_{n=1 \dots N-M}$, where $Y_n=\sum_{j=0 \dots M} a_{jn}\hat x_{n+j,u}$ by applying the left-hand side of Eq. \eqref{eq:var_step} to $\hat x_{n+j,u}$. 
Thus, it holds that $Y_n=\sum_{j=0}^M  b_{jm}\tilde f_u(\hat x_{n+j})+\epsilon_n$ with noise $\epsilon=(\epsilon_n)_{n=1 \dots N-M}$, corresponding $\lambda=(\lambda_n)_{n=1 \dots N-M}$ and $\lambda_n=\sum_{j=0}^M a^2_{jn}\sigma_u^2$.
For  $x^{\star} \in \mathbb R^d$, we obtain $K \in \mathbb{R}^{N-M \times N-M}$ and $k(x^{\star}) \in \mathbb{R}^{N-M}$, where 
\begin{equation}\label{eq:multistepkernels}
\begin{aligned}
(K)_{nm}& =\text{cov}\left(\sum_{j=0}^M b_{jn}\tilde f_u(\hat x_{n+j}),\sum_{j=0}^M  b_{jm}\tilde f_u(\hat x_{n+j})\right) \\
&=\sum_{i=0}^M \sum_{j=0}^M b_{jn}k(\hat x_{n+j},\hat x_{m+j}) b_{im}
\end{aligned}
\end{equation}
and
\begin{equation}
\begin{aligned} 
(k(x^{\star}))_n&=\text{cov}\left(\tilde f_u(x^{\star}),\sum_{j=0}^M b_{jn}\tilde f_u(\hat x_{n+j})\right) \\ &=\sum_{j=0}^M b_{jn} k(x^{\star},\hat x_{n+j}).
\end{aligned}
\end{equation}
\paragraph{Taylor integrators:}
We obtain observations $Y=(Y_n)_{n=1 \dots N-1}$, where $Y_{n}=\hat x_{n+1,u}-\hat x_{n,u}$
Thus it holds that $Y_n=\sum_{l=1}^P \frac{h_n^l}{l!}\tilde f_u^l(\hat x_n)+\epsilon_n$ with  $\epsilon=(\epsilon_n)_{n=1 \dots N-1}$ and corresponding $\lambda=2\sigma_u^2$ (cf. Eq. \eqref{eq:Taylor}).
In order to make inference tractable, we approximate each dimension $u$ of the truncated Taylor representation of order $P$ (cf. Eq. \eqref{eq:Taylor}) with $P$ independent GPs $\tilde f_u^l:\mathbb{R}^d \rightarrow \mathbb{R}$ and kernels $k_{l}$, where $l=1, \dots, P$.
Despite this approximation, we are able to derive error bounds.
With the independence assumption, we obtain for  $x^{\star} \in \mathbb{R}^d$ that $K \in \mathbb{R}^{N-1 \times N-1}$ and $k_i(x^{\star}) \in \mathbb{R}^{N-1}$, where
\begin{equation} \label{eq:Taylorkernel}
\begin{aligned}
(K)_{nm}&=\text{cov}\left(\sum_{l=1}^P \frac{h_n^l}{l!} \tilde f_u^l(\hat x_n),\sum_{l=1}^P \frac{h_m^l}{l!} \tilde f_u^l(\hat x_m)\right)\\
&=\sum_{l=1}^P\frac{h_n^l h_m^l}{l!l!} k_{l}(\hat x_n,\hat x_m)
\end{aligned}
\end{equation}
and
\begin{equation}
\begin{aligned}
(k_i(x^{\star}))_n &=\text{cov}\left(\tilde f_u^i(x^{\star}),\sum_{l=1}^P \frac{h_n^l}{l!} \tilde f_u^l(\hat x_n)\right)\\
&=\frac{h_n^i}{i!}k_{i}(x^{\star},\hat x_n).
\end{aligned}
\end{equation}
In our experiments, we train separate kernels and hyperparameters $k_{l}$, which is computationally efficient due to parallelization.
However, our framework also offers the option to compute adapted kernels.
Intuitively, the kernels and hyperparameters representing all higher-order derivatives depend on each other (see Appendix Sec. 1.2 for details). 

\paragraph{Training and inference: }
Analogous to the standard scenario, we parametrize the kernel functions $k$ and $k_l$ with $\theta$.
The trainable parameters $\theta$ and the noise parameter $\sigma_u$ are adapted by maximizing Eq. \eqref{eq:loglikeli}. 
Here, $K$ is obtained via Eq. \eqref{eq:multistepkernels} in the multistep and via Eq. \eqref{eq:Taylorkernel} in the Taylor case. 
Similarly, calculating the posterior distribution is now straightforward.
In the multistep case, we obtain the posterior distribution for $f_u(x^{\star})$ by applying  Eq. \eqref{eq:sparse-moments} to the corresponding transformed $Y$, $\lambda$ and the kernels in Eq. \eqref{eq:multistepkernels}.
In the Taylor case, we obtain the posterior distribution for $f_u^i(x^{\star})$ by applying Eq. \eqref{eq:sparse-moments} to the corresponding transformed $Y$, $\lambda$ and kernels in \eqref{eq:Taylorkernel}. Here, $k(x^{\star})$ in Eq. \eqref{eq:sparse-moments} is replaced by $k_i(x^{\star})$. 

\subsection{Sampling}
Standard GP sampling via Eq. \eqref{eq:sparse-moments} is computationally not tractable for sampling trajectories, being an iterative process. 
This is due to the fact that all dynamics evaluations have to be conditioned on previous ones.
To make trajectory sampling feasible, we derive a DS scheme similar to Eq. \eqref{eq:matheron} for multistep and Taylor integrators by applying Matheron's rule \citep{howarth1979} to the posterior. 

\paragraph{Multistep: }
We sample from $\tilde f_u$ via
\begin{equation}
\begin{aligned}
\tilde f_{u}(\cdot|Y) & \sim \sum_{k=1}^S w_k\phi_k(\cdot)+k(\cdot)(K+\lambda I)^{-1} \\ &\left(Y-\sum_{k=1}^S w_k F_k-\epsilon\right),
\end{aligned}
\end{equation}
with $F_k=(F_{k,n})_{n=1 \dots N} \in \mathbb{R}^{N-M}$ and $(\tilde F_{k,n})=\sum_{j=0}^M b_{jn}\phi_k(X_n)$.
Here, samples from the prior $\tilde f_u$ are represented by $S$ random Fourier bases $\phi_k$ and corresponding weights $w_k$ (cf. Eq. \eqref{eq:matheron}).
The time-varying noise is represented by $\epsilon \sim \mathcal{N}(0,\lambda I)$.
 
\paragraph{Taylor: }
We sample from the Taylor component $\tilde f_u^i$ via 
\begin{equation}
\begin{aligned}
\tilde f_u^i(\cdot|Y)& \sim \sum_{k=1}^S  w_{ki} \phi_{ki}(\cdot)+k_i(\cdot)(K+\lambda I)^{-1} \\
& \left(Y- \sum_{k=1}^S \sum_{l=1}^P w_{kl} \tilde F_{kl}-\epsilon \right),
\end{aligned}
\end{equation}
with $\tilde F_{kl}=(\tilde F_{kl,n})_{n=1 \dots N} \in \mathbb{R}^{N-M}$ and $\tilde F_{kl,n}=\frac{h_n^l}{l!} \phi_{kl}(X_n).$
We represent the prior $\tilde f_u^l, l=1, \dots, P$ via $S$ random Fourier bases $\phi_{kl}$ and weights $w_{kl}$ (cf. Eq. \eqref{eq:matheron}). 
The noise is represented by $\epsilon \sim \mathcal{N}(0,\lambda  I)$. 
We obtain the full truncated Taylor series for dimension $u$ by sampling from $\tilde f_u^l(\cdot|Y), l=1, \dots, P$. 
See Appendix Sec. 1.1.2 for detailed derivations of kernels and prediction schemes. 

\subsection{Predictions} 
The proposed DS schemes allow treating the trained model like a standard ODE since full dynamics functions can be sampled from the posterior and evaluated at arbitrary points.
In particular, we are able to integrate the learned model with arbitrary integrators including adaptive step size ones \citep{hairerVol1}.
These methods integrate system $\eqref{eq:dyn}$ with arbitrary accuracy leading to a negligible numerical error.
We can therefore leverage them to evaluate if our learned dynamics $\tilde f$ is indeed a good ODE model similar to \citep{ott}. To this end, the error is evaluated with an adaptive step-size integrator on a train or validation set indicating whether the model represents the ODE accurately. If not, it is retrained with higher accuracy.
In the experiments, we demonstrate that multistep integrators are suitable to learn continuous-time dynamics from regularly and irregularly-sampled grids. 
Taylor integrators, however, are especially suitable for varying step sizes. 
This is due to the fact that the Taylor representation \eqref{eq:Taylor} does not force the GPs $\tilde{f}^l_u $ to learn the correct part of the Taylor series since the step size has no influence on the learning task in the regularly-sampled case.
However, due to the uniqueness of the Taylor series, varying step sizes limit the freedom of the GPs.
Even if this is not the focus of this paper, our method is also useful to obtain structure-preserving predictions as addressed in \citep{https://doi.org/10.48550/arxiv.2102.01606}.
Embedding the physical structure (e.g. Hamiltonian) in the GP kernel and predicting the trained model with a structure-preserving integrator (e.g. a symplectic one) yields predictions that are for example volume-preserving. 

\subsection{Error Estimates for Multistep Methods}
We aim to quantify the accuracy of our learned dynamics by bounding $\Vert f_u(x)-\mu(x) \Vert$, where $u=1 \dots d$ and $x \in \mathbb{R}^d$ and $\mu(x)$ the mean approximation. Here, $f=(f_1,\dots,f_d)^T$ denotes the true ODE dynamics (cf. Eq. \ref{eq:dyn}), $\mu(x)=k(x)^T(K+\lambda I)^{-1}Y$ the posterior mean and $\sigma^2(x)=k(x,x)-k(x)^T(K+\lambda I)^{-1}k(x)$ the posterior variance, where $Y$, $K$ and $k(x^{\star})$ are calculated as described in Sec. \ref{sec:multistep}.
We consider noiseless data and a constant jitter $\lambda>0$.
We assume $f_u \in H_u$, where $H_u$ denotes a reproducing kernel Hilbert space (RKHS) represented by the kernel $k$. 
The main idea is to combine GP error bounds such as \citet{multiarm} or \citet{Fiedler_Scherer_Trimpe_2021} with the error of the integrators.
\begin{theorem}[Multistep error]
\label{thm}
	Consider a multistep method of order $P$  with coefficient matrices $A$ and $B$  (cf. Eq. \eqref{eq:var_step}).  
    Assume $f_u \in H_u$ with kernel $k$ and RKHS norm $\Vert f_u \Vert_{k} \leq C$. Further, assume that $|f_u^{P+1}|_{\Omega} \leq L$ and $|f_u^{P+2}I|_{\Omega} \leq L$. 
	Under mild assumptions and with $\lambda=1+\tau$ it holds
	\begin{equation}
	\begin{aligned}
	\Vert f_u(x)-\mu(x)\Vert \leq& \sigma(x)\sqrt{\Vert((K+\tau I)^{-1}+I)^{-1}\Vert} \\
	(C+&\mathrm{Constant}(N,M,\max(h),A,B,P,L,\lambda)).
	\end{aligned}
	\end{equation}	
\end{theorem}
Thm. \ref{thm} enables estimating the model error based on its uncertainty estimates.
Intuitively, a small GP uncertainty $\sigma(x)$ together with a small RKHS norm of the true dynamics and an accurate integrator yields accurate ODE dynamics.
We provide details, proofs, and a similar result for Taylor integrators in Appendix Sec. 2. 
 

%% file: related_work.tex
\section{Related Work}
Dynamics model learning is a broad field and has been addressed by different communities for many years \citep{dynRobot, ljung1999system}.
Most GP dynamics model learning approaches consider discrete-time systems \citep{doerr2018probabilistic, 10.5555/3104482.3104541}.
While discrete-time systems are typically modeled with one-step ahead predictions or history-based approaches, there are also approaches that apply numerical integrators \citep{https://doi.org/10.48550/arxiv.2102.01606, HamGP, pmlr-v168-brudigam22a}. 
However, in contrast to this work, they focus on the preservation of physical structure.

Continuous-time models are often learned with neural ODEs \cite{NIPS2018_7892}. 
To this end, the dynamics are modeled with a neural network and integrated with an integrator of desired accuracy. 
\citet{https://doi.org/10.48550/arxiv.2206.07335} provide error estimates for the accuracy of neural ODEs trained on Runge-Kutta integrators.
Neural networks have been trained on multistep integrators as well \citep{https://doi.org/10.48550/arxiv.1801.01236} followed by error analysis \citep{doi:10.1137/19M130981X}. 
The most recent work bounds the error based on inverse modified differential equations \citep{multisteperror}. 
However, none of these approaches leverages varying step size multistep methods.
Thus, the approaches and results can not be applied to irregularly-sampled data. 
\citet{Djeumou_2022} leverage Taylor integrators for neural ODE predictions.
However, in contrast to this work, they do not leverage them for training.
Further, all of the above approaches refer to deterministic neural networks and thus do not provide uncertainty estimates. 
Instead, we aim for a probabilistic GP model that can be trained and updated via exact inference and provides error bounds.
Multistep methods have been combined with GPs in \citet{10.5555/3157382.3157580, raissi2017numerical}.
However, they address the probabilistic numerics setting, where the dynamics are known. Instead, we consider the problem of inferring unknown dynamics from data. 

Some works combine ODE learning with GPs.
Gradient matching methods model the measured trajectories with a GP \citep{Wenketal19, Dondelinger13odeparameter}. However, they consider a parametric dynamics model with unknown parameters for the dynamics.
The concept is extended in \citet{heinonen2014learning} to nonparametric models. 
However, they still consider the gradient matching approximation and the learned dynamics are not a GP.
Some approaches also model ODE dynamics with GPs.  
\citet{DBLP:conf/icml/HeinonenYMIL18, pmlr-v180-hegde22a} tackle the problem by applying sparse GPs and train them similarly to neural ODEs. 
This requires an approximation via variational inference, while we aim for exact inference. 
\citet{DBLP:journals/corr/abs-2211-11103} consider standard GP conditioning as well by linearizing parts of the dynamics systems, which allows to propagate the uncertainty of the approximated system.
However, this requires approximations at many points and the uncertainty is not propagated exactly.

%% file: exp.tex
\section{Experiments}
\label{sec:exp}

Next, we evaluate our methods numerically and support the intuitive and theoretical findings. 
Our framework provides the option to train with various integrators of arbitrary order and time irregularity. 
In contrast to most GP dynamics models, we treat the learned dynamics indeed as an ODE and perform predictions with an integrator that solves the ODE almost exactly.
In particular, the prediction integrator does not necessarily correspond to the training integrator. 
We consider both, mean and DS predictions and conduct experiments with fixed and varying step sizes. 
We show that
(i) for multistep integrators, the higher the order, the better the ODE approximation on regular and irregular grids. 
(ii) Taylor integrators are especially effective on irregular grids (cf. Sec \ref{sec:multistep}).
(iii) We can compete with a variational inference baseline.
(iv) Through extensive experiments, we investigate which integrator to use in which scenario including their limitations. We further demonstrate that our framework can cope with different choices of integrators. 
\paragraph{Integrators: }
We consider the three main classes of multistep integrators:
Adam-Bashforth (AB) methods, representing explicit integrators; and Adam Moulton (AM) and backward-difference formulas (BDF), both representing implicit integrators that require the solution of a minimization problem (see \citet{hairerVol1}).
For all integrators, we consider orders (of consistency) 1 to 3. 
We refer to them as "integrator"+"order".
So, Adam-Bashforth of order 1 would be denoted "AB 1".
Since Taylor 1 and AB 1 correspond to the explicit Euler, we conduct these experiments only once. 
Similarly, AM 1 and BDF 1 correspond to the implicit Euler method. 
For predictions, we consider the Runge-Kutta 4(5) integrator, an adaptive step size integrator that makes the numerical error negligible and solves the ODE \eqref{eq:dyn} almost exactly \citep{hairerVol1}.
We refer to it as RK4(5).
This allows to quantify whether our model provides an accurate ODE approximation.
In Appendix Sec. 3.2, we provide results for training and predictions with identical integrator.
On regular grids, this provides comparable results for all integrators.
Thus, the differences in accuracy are caused by the ODE approximation qualities. 
\paragraph{Experimental setup: }
For all experiments, we consider DS predictions by drawing independent trajectories from the GP posterior and computing the statistical mean and variance. 
We evaluate the mean squared error (MSE) between data and predictions on five independent runs and report mean and standard deviation.
We consider the explicit Euler (AB 1) as a baseline.
We further compare to the GP-ODE proposed in \citet{pmlr-v180-hegde22a}, a variational inference-based approach that works similarly as a neural ODE.
By considering orders 1 to 3 for each integrator, we investigate how the order affects the results.
All GPs are modeled with ARD kernels \citep{10.5555/1162254}.
We consider simulated systems and real-world data. 
In Appendix Sec. 3 and 4, we report details, runtimes and additional results.
This includes results for the Taylor integrators with adapted kernels (cf. Sec. \ref{sec:multistep}). 
\subsection{Systems}
Next, we specify the systems and learning tasks. 
Details for the simulated systems are provided in Appendix Sec. 4.1. 

\paragraph{Damped harmonic oscillator (DHO): }
The DHO system represents an oscillator subject to a damping or friction force \citep{multisteperror}. 
We consider a timeline with regular step size and generate a trajectory of 10 seconds and step size $h=0.01$.
The first 500 steps are used for training, while predictions are performed on the full trajectory. 
We consider the multistep integrators AB, AM and BDF, order 1 to 3.
We investigate if indeed the accuracy of the ODE increases with the order. 
The results are displayed in Table \ref{DHO}, an example for DS predictions is Fig. \ref{fig:nsep} (left). 
\paragraph{Van-der-Pol oscillator (VDP): } 
We simulate the VDP system \citep{cveticanin2013} on an irregular timeline by sampling the step size within a certain range $b$ via
$t_{i + 1} = t_i + h( 1 + (w - 1/2)b),\text{ with } w \in \mathcal{U}(0, 1)$ and step size $h=0.1$. 
Here, we choose $b=0.5$. In Appendix Sec. 3.3, we add results for $b=0.3$.
We compute rollouts with 100 steps.
Training is performed on the first 50 steps, predictions are performed on the full trajectory.
We consider AB, AM and BDF as well as Taylor, order 1 to 3.
The results are displayed in Table \ref{VDP}, DS rollouts and phase plots in Fig. \ref{fig:sep}.

\paragraph{Real spring system: }
We consider measurements from a linear mass-spring system on an air track from \citet{distillinglaws}. 
The dataset corresponds to a 871 time steps long trajectory where each point is a vector $\text{x} = (x,v)$ of position and velocity.  
We use the first 400 steps for training, while predictions are performed on the full trajectory.
We consider AB, AM and BDF.  
The results are displayed in Table \ref{real}, an example for DS rollouts including GP uncertainty in Fig. \ref{fig:nsep} (middle). 

\paragraph{Human motion data (MoCap): }
Like \citet{pmlr-v180-hegde22a}, we consider experimental human motion data from CMU MoCap database for subject 09 short.
We also use the same trajectories for training and testing as them.
The 50-dimensional data are projected into a 3-dimensional space by applying a PCA.
We learn the ODE in the PCA space and project it back to the original space to obtain predictions.  
This results in some loss of information.
However, since we aim to apply exact inference, we can not embed our method in a latent space in contrast to \citet{pmlr-v180-hegde22a}.
The results are displayed in Table \ref{Mocap}, a mean rollout in Fig. \ref{fig:nsep} (right). 
\subsection{Results}
\begin{figure*}[tb]
	\centering
	\begin{minipage}[htb]{0.3\textwidth}
		\centering
		\includegraphics[width= \textwidth]{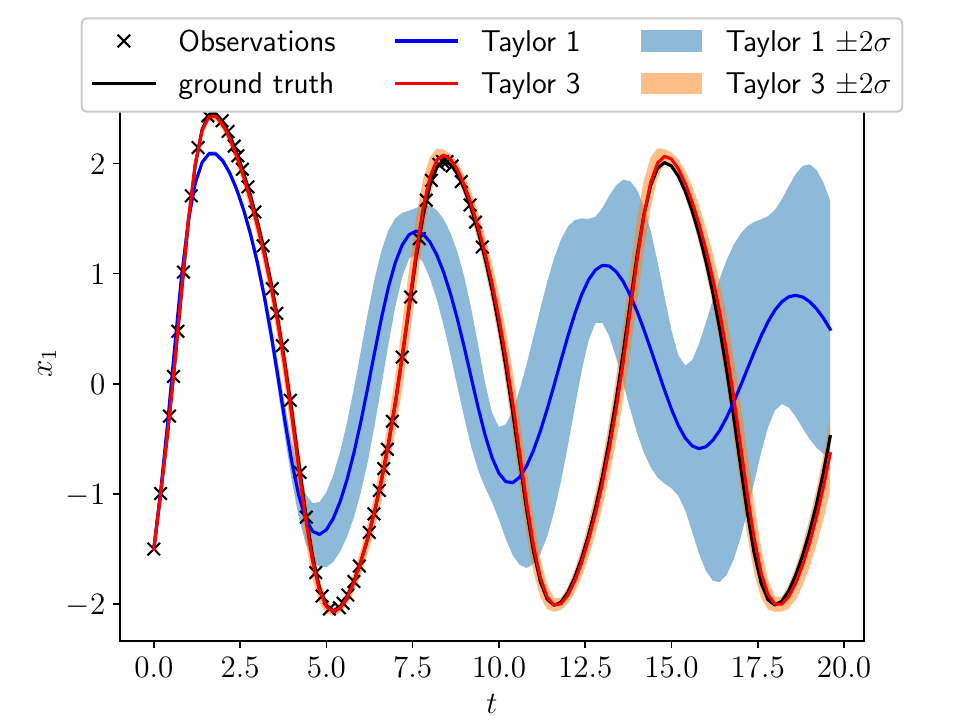}
	\end{minipage}
	\begin{minipage}[htb]{0.3\textwidth}
		\centering		\includegraphics[width= \textwidth]{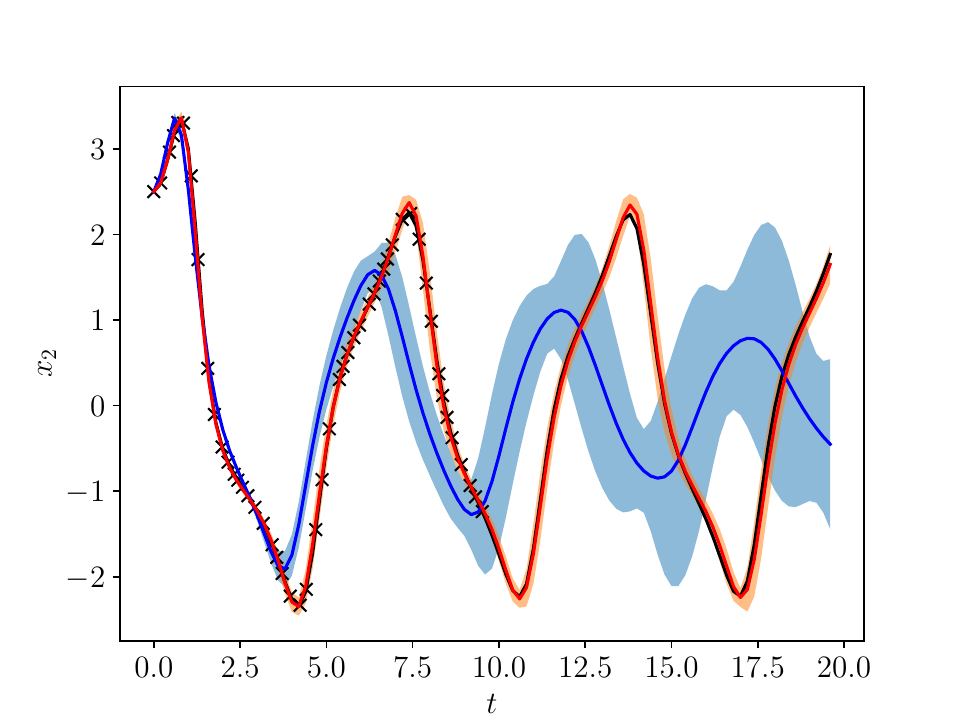}
	\end{minipage}	
	\begin{minipage}[htb]{0.3\textwidth}
		\centering
		\includegraphics[width= \textwidth]{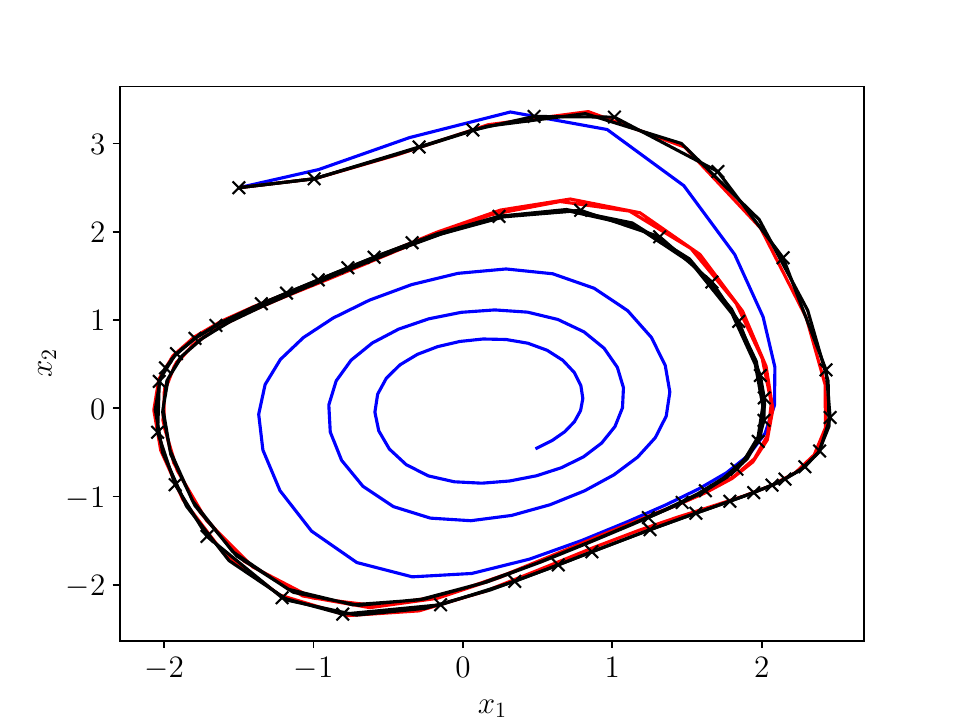}
	\end{minipage}	
	\caption{DS predictions (left, middle) and corresponding phase (right) for the VDP system with Taylor order 1 and 3. Shaded regions indicate the GP uncertainty. With increasing order, a clear improvement is visible.}
	\label{fig:sep}
\end{figure*}

\begin{figure*} 
		\centering 
	\begin{minipage}[htb]{0.3\textwidth}
		\centering
		\includegraphics[width= \textwidth]{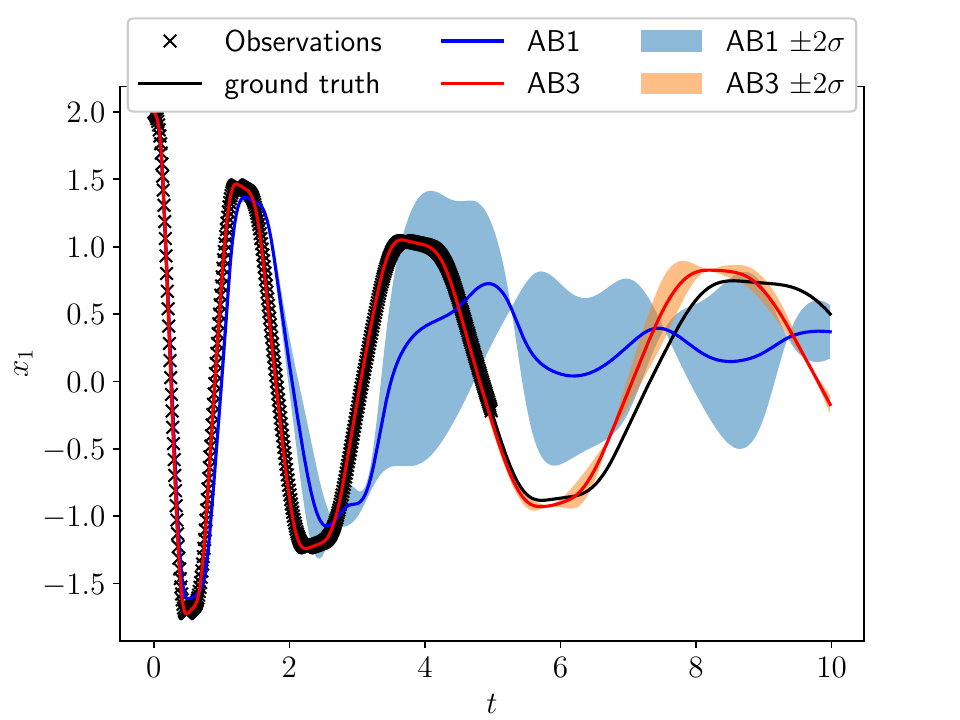}
	\end{minipage}
	\begin{minipage}[htb]{0.3\textwidth}
		\centering		\includegraphics[width= \textwidth]{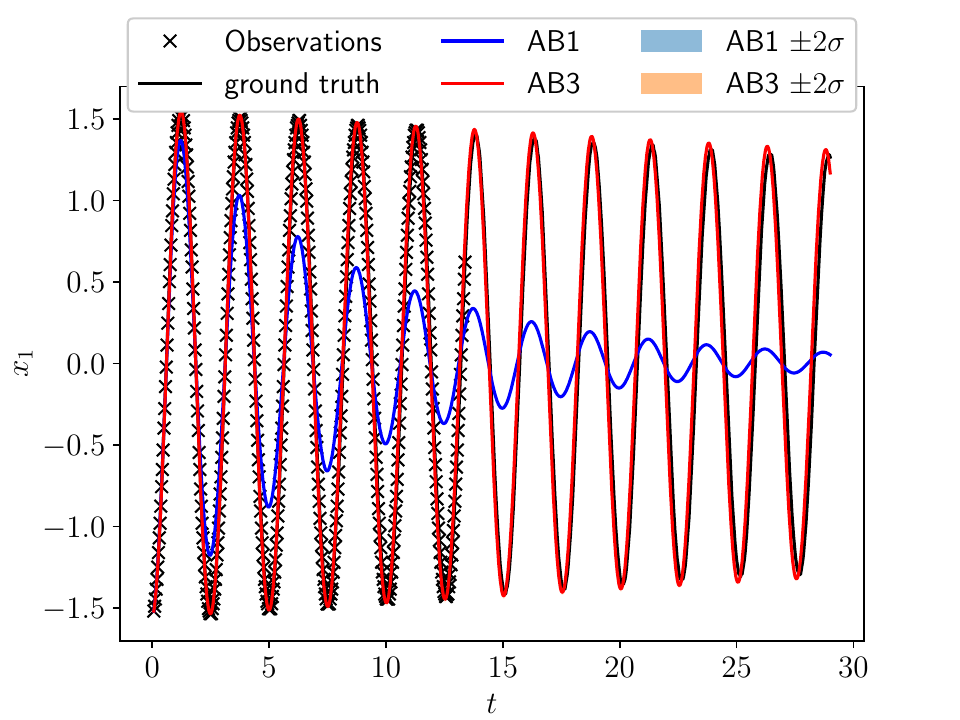}
	\end{minipage}	
	\begin{minipage}[htb]{0.3\textwidth}
		\centering
		\includegraphics[width= \textwidth]{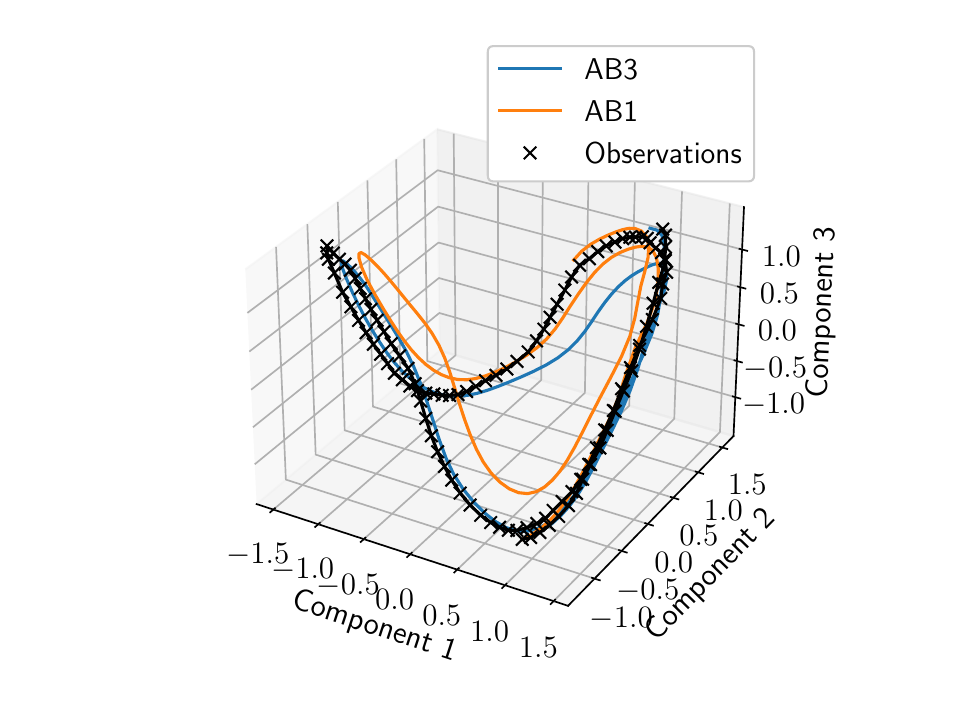}
	\end{minipage}	
	\caption{DS predictions on a single seed for the DHO system (left) and the real spring system (middle). Uncertainty is illustrated via shaded regions. Mean predictions in PCA space for MoCap (right). Raising the integrator order yields an improvement.}
	\label{fig:nsep}
\end{figure*}
The results demonstrate that accurate ODE models from regular and irregular timelines can be obtained with higher-order integrators while maintaining the advantages of exact GP inference.
This also means that the standard learning procedure via explicit Euler (AB 1) does not provide good ODE representations.
Further, the results coincide with the theoretical findings and the intuition, i.e. the accuracy increases with the order.
However, the approach has limits if the step sizes are too large or too irregular. 
In practice, the necessary order to obtain an accurate ODE has to be evaluated numerically, e.g. by computing the error with RK4(5) on a validation set (cf. Sec. \ref{method}).
\textbf{DHO: } The results for the DHO system demonstrate that higher-order multistep methods allow learning an ODE that outperforms the GP-ODE in terms of accuracy. 
Further, the accuracy increases with the order (cf. Fig \ref{fig:sep} (left)). 
While the implicit integrators (AM and BDF) provide accurate results already for order 2, the explicit method (AB) struggles up to order 3. 
After entering the extrapolation area, the higher-order integrators deviate from the trajectory as well after some time (cf. Fig \ref{fig:sep} (left)).
This behavior is even stronger for the GP-ODE baseline.
However, AB 1 already deviates in the training area. 
\textbf{VDP: } The results for VDP extend these findings to the irregularly-sampled case (cf. Fig. \ref{fig:nsep}).
In contrast to the DHO system, the implicit BDF method still causes problems for order 2.
Further, Taylor methods work well for varying step size as described in Sec. \ref{sec:multistep}.
In Appendix 3.2, we demonstrate that here, higher order yields higher accuracy also if we predict with the training integrator. 
This confirmes that the explicit Euler is not suitable for varying step sizes. 
\textbf{Real-world data: }
The findings for the real spring system methods show that we can also learn accurate ODEs from real-world data (cf. Fig \ref{fig:sep} (middle)).
On the MoCap data, our integrators are outperformed by the GP-ODE when sampling with DS.
Also, increasing accuracy with the order is not clearly visible for DS predictions.
This might be caused by the high observation noise our method learns.
Further, due to the variational inference setting, the GP-ODE can be trained on the latent space balancing information loss caused by the PCA.
Thus, the two methods are not fully comparable.
In contrast, our mean predictions are accurate and the typical trend is visible (cf. Fig \ref{fig:sep} (right)).

\begin{table}[!ht] 
	\centering
	\begin{tabular}{rcc}
		\hline
	    Integrator & order & MSE  \\ \hline
   		Baseline GP-ODE  & - & 0.187 (0.091) \\   
		Baseline AB & 1 &  0.750 (0.232)\\ 
		AB & 2 &  2.62 (0.011) \\ 
		AB & 3 &  0.062 (0.027) \\ 
		AM/ BDF & 1 &  1.175 (0.060)\\ 
		AM & 2 &  0.027 (0.015)\\ 
		AM & 3 &  0.043 (0.023)\\ 
		BDF & 2 &  0.009 (0.006)  \\ 
		BDF &3 &  \textbf{0.006} (0.001) \\

	\end{tabular}
\caption{DS predictions with RK4(5) for DHO}
\label{DHO}
\end{table}

\begin{table}[!ht]
	\centering
	\begin{tabular}{rcc}
		\hline
		Integrator & order & MSE   \\ \hline
        Baseline GP-ODE & - &  0.043 (0.028)\\
		Baseline AB & 1 & 1.843 (0.141)  \\ 
		AB & 2  & 0.030 (0.007) \\ 
		AB & 3  & 0.114 (0.058)  \\ 
		AM/BDF & 1  & 34.578 (18.372) \\ 
		AM &2  & 0.015 (0.003) \\ 
		AM &3  & 0.016 (0.005)  \\ 
		BDF & 2 & 17.753 (30.677) \\ 
		BDF & 3  & 0.063 (0.037)\\
		Taylor & 2 & 0.01 (0.004)  \\  
        Taylor & 3 & \textbf{0.005} (0.002) \\   

	\end{tabular}
\caption{DS predictions with RK4(5) for VDP}
\label{VDP}
\end{table}

\begin{table}[!ht] 
	\centering
	\begin{tabular}{rcc}
		\hline
		Integrator & order &  MSE   \\ \hline
		Baseline GP-ODE & - & 0.163 (0.093) \\
		Baseline AB & 1  & 0.502 (0.029)  \\  
		AB & 2 &  0.026 (0.037) \\ 
		AB & 3 &  \textbf{0.007} (0.009) \\ 
		AM/ BDF & 1  & 420.5 (422) \\  
		AM & 2  & 0.014 (0.004)  \\  
		AM & 3  & 0.016 (0.014) \\  
		BDF & 2 & 0.050 (0.037)  \\ 
		BDF & 3 & 0.035 (0.041)\\  

	\end{tabular}
\caption{DS predictions with RK4(5) for real spring system}
\label{real}
\end{table}

\begin{table}[!ht] 
	\centering
	\begin{tabular}{rccc}
		\hline
		Integrator & order & type & MSE   \\ \hline
    	Baseline GP-ODE & - & DS & \textbf{20.71} (1.25) \\  
		Baseline AB & 1  & DS & 40.97 (15.47) \\ 
		AB & 2 & DS & 55.79 (46.44) \\ 
		AB & 3 & DS & 30.42 (9.20) \\ \hline 
	    Baseline AB & 1  & mean & 37.76 (1.39) \\ 
		AB & 2 & mean & \textbf{10.12} (0.33) \\ 
		AB & 3 & mean & 11.78 (1.58) \\ 

	\end{tabular}
	\caption{Mean and DS predictions with RK4(5) for MoCap}
	\label{Mocap}  
\end{table}

%
%

%% file: conclusion.tex
\section{Conclusion and Future Work}
\label{sec:conclusion}
We propose a flexible framework to learn ODEs via exact GP inference from discrete data for both, equidistant and variable time steps.
To this end, we leverage multistep and Taylor integrators. 
Exact inference has several advantages compared to approximations such as variational inference, e.g., it provides mathematical guarantees.
In order to allow the usage of arbitrary integrators for predictions, we derive decoupled sampling schemes.
This enables sampling consistent vector fields from the GP posterior.
We show theoretically and empirically that integrators of sufficient order provide accurate ODE approximations.
Interesting aspects for future work include the application of such schemes, e.g., to model-based reinforcement learning.

%% file: document.tex
\section{Mathematical details}\label{sec:der}
In this section, we provide more detailed derivations of inference, noise calculation and sampling schemes. 
We further demonstrate how matching function spaces can be computed for Taylor kernels, where the dynamics are modeled with ARD kernels.
We adopt the notations from the main paper and start with a detailed derivation of the noise. 

\subsection{Noise calculation}
As explained in the method section, we obtain adapted GP observations, since linear combinations of trajectory points correspond to linear combinations of dynamics evaluations. 
Therefore, the corresponding noise is time-varying for irregularly-sampled data and correlated between different observations. 
In our framework, we provide three options: training with correlated and time-varying 
noise that we will derive in the following; training with a diagonal noise covariance matrix with time-varying entries by considering only the variances; and training with a diagonal matrix with constant entries (i.i.d noise). 
In our experiment, we choose the second option due to efficiency and numerical stability during training. 
\paragraph{Multistep integrators: }
In the multistep case, we obtain for dimension $u$ that  
\begin{equation} \label{eq:multiste_data}
\begin{aligned}
Y_n=\sum_{j=0}^M a_{jn}\hat x_{n+j,u}=\sum_{j=0}^M a_{jn} x_{n+j,u}+\epsilon_n,
\end{aligned}
\end{equation}
where $\epsilon=(\epsilon_n)_{n=1 \dots N-M}$ and $\epsilon_n=\sum_{j=0}^M a_{jn} \nu_{n+j,u}$.
Eq. \eqref{eq:multiste_data} demonstrates that different observations share parts of the observation noise.
It holds that $\epsilon \sim \mathcal{N}(0,\Sigma_y)$, where $\Sigma_y \in \mathbb{R}^{N-M \times N-M}$, $\Sigma_y= \sigma_u^2 \tilde A \tilde A^T$ and $\sigma_u^2$ denotes the observation noise. 
Here, the matrix $\tilde A \in \mathbb{R}^{N-M \times N}$ is constructed such that \begin{equation}
(\tilde A)_{n,m}=\begin{cases}
a_{m-n, n} & 0 \leq m-n \leq M \\
0 & \text{else}
\end{cases}
\end{equation}
\paragraph{Taylor integrators: }
In the Taylor case, we obtain data $Y=(Y_n)_{n=1 \dots N-1}$ for dimension $u$ that fulfill
\begin{equation} \label{eq:Taylor_data}
\begin{aligned}
Y_n=\hat x_{n+1,u}-\hat x_{n,u}=x_{n+1,u}-x_{n,u}+\epsilon_n,
\end{aligned}
\end{equation}
where $\epsilon=(\epsilon_n)_{n=1 \dots N-1}$ and $\epsilon_n=\nu_{n+1,u}-\nu_{n,u}$.
Again, we obtain correlated noise. 
This yields a noise matrix $\Sigma_y \in \mathbb R^{N-1 \times N-1}$, where \begin{equation}
(\Sigma_y)_{n,m}=\begin{cases}
2 \sigma_u^2 & n=m \\
-\sigma_u^2 &  |n-m|=1\\
0 & \text{ else }
\end{cases}.
\end{equation}

\subsection{Inference and sampling schemes}
In order to derive the equations for decoupled sampling, we recall Matheron's rule \citep{howarth1979}.
\begin{theorem}[Matheron's rule]
Let $a$ and $b$ be jointly Gaussian random variables. 
Then the random variable $a$ conditioned on $b=\beta$ is equal in distribution to 
\begin{equation}
(a|b=\beta)\sim a+  \textrm{cov}(a,b)  \textrm{cov}(b,b)^{-1}(\beta-b). 
\end{equation}
\end{theorem}
Next, we will describe in detail how an inference scheme can be derived for the Taylor method and how sampling schemes can be derived for the multistep and Taylor integrators. 
 
\paragraph{Multistep methods: }
In order to derive the decoupled sampling scheme for multistep methods, Matheron's rule is applied to the posterior conditioned on $Y$.
It holds that 
\begin{equation}
\begin{aligned}
& \tilde f_u(\cdot|Y)\sim \underbrace{\tilde f_{u}(\cdot)}_{\text{prior}}+k(\cdot)(K+\Sigma_y)^{-1} \left(Y-Z -\epsilon \right) \\
&\approx  \sum_{l=1}^S w_l \phi_l(\cdot) +k(\cdot)(K+\Sigma_y)^{-1}(Y-Z-\epsilon),
\end{aligned}
\end{equation}
where $Z=(Z_n)_{n=1 \dots N-M}$ and
\begin{equation}
Z_n=\sum_{j=0}^{M}b_{jn} \sum_{l=1}^S w_l \phi_l(X_{n+j}).
\end{equation}
Here, we represent samples from the prior $\tilde f_{u}(\cdot)$ by $S$ Random Fourier bases $\phi_l$ and corresponding weights $w_l$.
Further, $\epsilon \sim \mathcal{N}(0,\Sigma_y)$ represents the time-varying noise (cf. Eq. \eqref{eq:multiste_data}).

\paragraph{Taylor integrator: }
Here, we derive the sampling schemes for Taylor integrators. 
Again, we focus on dimension $u$.
Applying Matheron's rule to the posterior allows us to sample from $\tilde f^l$ for each component $l=1,\dots,P$ of the truncated Taylor series.
This yields for dimension $u$
\begin{equation}
\begin{aligned}
&\tilde f^l_u(\cdot|Y_u)\sim \underbrace{\tilde f^l_u(\cdot)}_{\text{prior}} + k_l(\cdot)(K+\Sigma_y)^{-1} \left(Y-Z-\epsilon \right) \\
&=\sum_{j=0}^T  w_{lj}\phi_{lj}(\cdot)+k_l(\cdot)(K+\Sigma_y)^{-1}\left(Y-Z-\epsilon \right),
\end{aligned}
\end{equation}
where $Z=(Z_n)_{n=1 \dots N-1}$ with
\begin{equation}
Z_n=\sum_{i=1}^P \frac{h_{n}^i}{i!} w_{ij}\phi_{ij}(X_n).
\end{equation}
Here, we represent samples from the prior $\tilde f^i_u$ by $S$ Random Fourier bases $\phi_{ij}$, corresponding weights $w_{ij}$ and time-varying noise $\epsilon \sim \mathcal{N}(0,\Sigma_y)$ (cf. Eq. \eqref{eq:Taylor_data}).

\subsection{Deriving the right kernel for Taylor series}
In this section, we describe how to derive matching kernels for the Taylor schemes.  
Intuitively, we ensure that the kernels and hyperparameters representing the Lie-derivatives are adapted to each other. 
For the ARD kernel, we compute the kernels explicitly up to order 3.  
To this end, we recall the Taylor series 
\begin{equation}\label{eq:Taylor}
x(t+h)=x(t) +\sum_{l=1}^{\infty} \frac{h^l}{l!}f^l(x(t)).
\end{equation}
Asssume that dimension $j$ of the Lie-derivative $f_j^l: \mathbb R^{d} \rightarrow \mathbb R$ lies in a function space that can be represented by the kernel function $k_j^l$ for $j=1,\dots,d$. 
Then, the Lie derivative $f_u^{l+1}$ lies in a function space that can be represented by the kernel function $k^{l+1}_u$.
It holds that 
\begin{equation}\label{eq:Lie_kernel}
k^{l+1}_u = \sum_{j=1}^d \frac{d}{dx_j}\frac{d}{dy_j}\left[k_u^l(x,y)\right]k_1^j(x,y).
\end{equation}
A detailed derivation of Eq. \eqref{eq:Lie_kernel} is provided in Thm. \ref{thm:lie}.

\subsubsection{Higher order ARD kernels}
We apply the findings to ARD kernels.
Consider the ARD kernel $k_1^i(x,y)=\sigma_i^2 \exp \left( -\sum_{j=1}^d\frac{(x_j-y_j)^2}{2 l_{ij}^2} \right)$.
Applying Eq. \eqref{eq:Lie_kernel} yields  
\begin{equation}
k_2^i= \sigma_i^2 \sum_{j=1}^d \frac{\sigma_j^2}{l_{ij}^4} (l_{ij}^2-(x_j-y_j)^2) \exp\left(-\sum_{k=1}^d\frac{(x_k-y_k)^2}{2 \tilde l_{ij,k}^2}\right),
\end{equation}
where $\tilde{l}_{ij}^2=\frac{1}{1/l_i^2+1/l_j^2}$.

In order to compute $k_3^i$, we first differentiate $k_2^i$. 
For the computation of $\frac{d}{dx_l}\frac{d}{dy_l}\left[\frac{\sigma_j^2}{l_{ij}^4} (l_{ij}^2-(x_i-x_j)^2) \exp \left(-\sum_{k=1}^d\frac{(x_j-y_j)^2}{2 \tilde l_{ij,k}^2}\right)\right]$, we have to distinguish into the case $j=l$ and $j \neq l$. 
Consider the case $l=j$. It holds that 
\begin{equation}\label{eq:x}
\begin{aligned}
&\frac{d}{dx_l}\left[(l_{il}^2-(x_l-y_l)^2) \exp\left(-\sum_{k=1}^d\frac{(x_k-y_k)^2}{2 \tilde l_{il,k}^2}\right)\right] \\
=&\left(\frac{1}{\tilde l_{il,l}^2}(x_l-y_l)^3-\left(2+\frac{l_{il}^2}{\tilde l_{il,l}^2}\right)(x_l-y_l)\right)\exp\left(-\sum_{k=1}^d\frac{(x_k-y_k)^2}{2 \tilde l_{il,k}^2}\right).
\end{aligned}
\end{equation}
For $l \neq j$ it holds that 
\begin{equation}
\begin{aligned}
&\frac{d}{dx_l}\left[(l_{ij}^2-(x_j-y_j)^2) \exp\left(-\sum_{k=1}^d\frac{(x_k-y_k)^2}{2 \tilde l_{ij,k}^2}\right)\right]\\
=&-\frac{1}{\tilde l_{ij,l}^2}(l_{ij}^2-(x_j-y_j)^2)(x_l-y_l)\exp\left(-\sum_{k=1}^d\frac{(x_k-y_k)^2}{2\tilde l_{ij,k}^2}\right).
\end{aligned}
\end{equation}
By differentiating also for $y_l$ and combining the two cases again, we obtain 
\begin{equation}
\begin{aligned}
\frac{d}{dx_l}&\frac{d}{dy_l}k_2^i=\sigma_i^2 \Bigg[\sum_{l=1}^d P_l^1 \exp\left(-\sum_{k=1}^d\frac{(x_k-y_k)^2}{2 \tilde l_{il,k}^2}\right)+P_l^2 \exp \left(-\sum_{k=1}^d\frac{(x_k-y_k)^2}{2 \tilde l_{ij,k}^2}\right)\Bigg],
\end{aligned}
\end{equation}
where
\begin{equation}
P_l^1=\frac{\sigma_l^2}{l_{il}^4}\left(\frac{5}{\tilde l_{il,l}^2}+\frac{l_{il}^2}{\tilde l_{il,l}^4}\right)\left(-(x_l-y_l)^2+\frac{1}{\tilde l_{il,l}^4}(x_l-y_l)^4+2+\frac{l_{il}^2}{\tilde l_{il,l}^2}\right)
\end{equation}
and
\begin{equation}
P_l^2=\sum_{j=1,j\neq l}^d  \frac{\sigma_j^2}{l_{ij}^4}\frac{1}{\tilde l_{ij,l}^4}(\tilde l^2_{ij,l}-(x_l-y_l)^2)(l_{ij}^2-(x_j-y_j)^2).
\end{equation}
Thus, for the kernel $k_3(x,y)$ it holds 
\begin{equation}
\begin{aligned}
k_3^i(x,y)& =\sum_{l=1}^d \frac{d}{dx_l}\frac{d}{dy_l}\left[k_2^i(x,y)\right]k_1^l(x,y) \\
=\sigma_i^2&\Bigg[
\sum_{l=1}^d P_l^3 \exp\left(-\sum_{k=1}^d\frac{(x_k-y_k)^2}{2 \tilde l_{ill,k}^2}\right) +P_l^4 \exp\left(-\sum_{k=1}^d\frac{(x_k-y_k)^2}{2 \tilde l_{ijl,k}^2}\right)
\Bigg],
\end{aligned}
\end{equation}
with $\tilde l_{ijl}^2=\frac{1}{1/l_i^2+1/l_j^2+1/l_l^2}$, 
\begin{equation}
P_l^3=\frac{\sigma_l^2}{l_{il}^4}\left(\frac{1}{\tilde l_{il,l}^4}(x_l-y_l)^4-\left(\frac{5}{\tilde l_{il,l}^2}+\frac{l_{il}^2}{\tilde l_{il,l}^4}\right)(x_l-y_l)^2+2+\frac{l_{il}^2}{\tilde l_{il,l}^2}\right)
\end{equation}
and 
\begin{equation}
P_l^4=\sum_{j=1,j\neq l}^d  \frac{\sigma_j^2}{l_{ij}^4}\frac{1}{\tilde l_{ij,l}^4}(\tilde l^2_{ij,l}-(x_l-y_l)^2)(l_{ij}^2-(x_j-y_j)^2).
\end{equation}

%% file: proofs.tex
\section{Proofs}\label{sec:proofs}
In this section, we derive bounds for our GP approximations. 
In particular, we derive expressions for the error
\begin{equation}
|\mu(x)-f_u(x)|, 
\end{equation}
where $\mu$ denotes the GP mean derived in the method section of the main paper and $f=(f_1 \dots f_d)$. 
The section is structured as follows: We first derive error bounds for the irregularly-sampled multistep methods presented in the method section of the main paper. 
Then, we derive similar error bounds for the irregularly-sampled Taylor integrators.
All proofs heavily rely on Lie derivatives, thus we first introduce the notation.
Further, the results assume correct choice of the underlying reproducing kernel Hilbert space (RKHS) and thus, the correct kernel.
This is usually not the case in practice since the kernel hyperparameters are optimized during training.
However, our results can be extended to mismatch of kernels by leveraging the results in  \citet{Fiedler_Scherer_Trimpe_2021}.

\subsection{Taylor representation} 
In this section, we assume that the the solution $x$ of the ODE $\dot{x}(t)=f(x(t))$ can be represented by a Taylor series via 
\begin{equation} \label{eq:Lie}
x(t+h)=\sum_{k=0}^{\infty}\frac{t^k}{k!}f^k(x),
\end{equation}
where $f^k: \mathbb R^d \rightarrow \mathbb R$ and $f^1=f$, $f^2=f^{\prime}f$, \dots, $f^{k+1}=\frac{d}{dx}f^kf$.
We denote the component-wise evaluation $f^k_u:\mathbb R^d \rightarrow \mathbb R$ for $u \in \{1 \dots d\}$ and $f^k=(f_1^k \dots f_d^k)$.
Next, we demonstrate how the corresponding reproducing kernel Hilbert spaces of the expansion $\eqref{eq:Lie}$ can be derived. 
\begin{theorem}[Gaussian processes on Lie-Derivatives] \label{thm:lie}
	Consider the Taylor representation of the autonomuous system \eqref{eq:Taylor} and assume that the dynamics $f_i(x)=f_i^1(x) \in H_i^1$, where $H_i^1$ denotes an RKHS with arbitrary smooth kernel $k^1_i(x,y)$ for $i=1,\dots,d$. 
	Then it holds that $f^l_i \in H_i^l$ with kernel $k_i^l$, where $k_i^l$ is defined via the following recursion
	\begin{equation}
	k^{l+1}_i(x,y)=\sum_{j=1}^d \frac{d}{dx_j}\frac{d}{dy_j}\left[k^l_i(x,y)\right]k^1_j(x,y),
	\end{equation}
	where $i=1,\dots,d$ denotes dimension $i$.
\end{theorem}
\begin{proof}
	The case $l=1$ is clear.
	Assume that the statement holds for $l$.
	Then, for $H^{l+1}_i$, it holds that $f_i^{l+1}=\sum_{j=1}^d \frac{d}{dx_j}\left[f_i^{l+1}(x)\right]f_j(x)$.
	Since $k_i^1$ is arbitary smooth, $\frac{d}{dx_j}\left[f_i^{l+1}(x)\right]$ can be represented in an RKHS 
	with kernel $k^{l+1}_i(x,y)=\frac{d}{dx_j}\frac{d}{dy_j}k^l_i(x,y)$ \citep{ZHOU2008456}.
	Since, $f_i \in H_i^1$, it holds that $\frac{d}{dx_j}\left[f_i^{l+1}(x)\right]f_j(x)$ can be represented with kernel 
	$\frac{d}{dx_j}\frac{d}{dy_j}k^l_i(x,y) \cdot k^1_j(x,y)$, where we applied \citet{aronszajn50reproducing} Ch. 8, Theorem 1. 
	Thus, the result follows.
\end{proof}
\subsection{Multistep methods}
Here, we present results for the multistep methods on irregularly-sampled step sizes.
The main idea is to first derive GP error bounds for the multistep kernels.
Afterwards, they are combined with the error of the integrators. 
We start with the error bounds for multistep kernels. 

\begin{theorem}[Multistep approximation]\label{multistepGP}
	Let $f: \mathbb{R}^d \rightarrow \mathbb{R}^d$. Here, we consider the dimension-wise evaluation $f_u:\mathbb{R}^d \rightarrow \mathbb{R} \in H_u$, where $H_u$ denotes an RKHS with corresponding kernel function $k$ and $u \in \{1\dots d\}$. 
	Consider $N$ training inputs $X \in \mathbb{R}^{d \times N}$ with $X=(X_n)_{n=1 \dots N}$ and $X_n \in \mathbb{R}^d$.
	Further, consider slightly pertubed observations $Y=\tilde Y+r \in \mathbb{R}^{N-M}$,
	with $Y=(Y_n)_{n=1 \dots N-M} \in \mathbb R^{N-M}$, $\tilde Y=(\tilde Y_n)_{n=1 \dots N-M} \in \mathbb R^{N-M}$, and $r=(r_n)_{n=1 \dots N-M} \in \mathbb R^{N-M}$, where 
	\begin{equation}
	\tilde Y_n=\sum_{j=0}^M b_{n+j} f_u(X_{n+j}).
	\end{equation}
	Further, assume
	\begin{equation}
	|Y-\tilde Y| \leq \epsilon.
	\end{equation}
	Here $B=(b_{jn})_{j=1,\dots,M}^{n=1,\dots,N}$ denotes the coefficients of an M-step multistep method with varying step sizes. 
	Further assume that $\Vert f_u \Vert_{k} \leq C$. 
	Then, for the GP mean approximation $\mu(x)=k(x)(K+\lambda I)^{-1}Y$ with $K \in \mathbb{R}^{N-M \times N-M}$ and $(K)_{nm}=\sum_{i=0}^M \sum_{j=0}^M b_{jn}k(X_{n+j},X_{m+j})b_{im}$, $k(x)\in\mathbb{R}^{N-M}$ with $k(x)=\sum_{j=0}^M b_{jn} k(x^{\star},X_{n+j})$, it holds that 
	\begin{equation}
	|\mu(x)-f_u(x)|\leq \sigma(x)\left(C+(1+\tau)^{-\frac{1}{2}}\epsilon \sqrt{\Vert((K+\tau I)^{-1}+I)^{-1} \Vert_2} \right)
	\end{equation}   	
	with $\lambda=1+\tau$.
\end{theorem}

\begin{proof}
	The proof is analogous to \citet{multiarm} in most parts.
    Define $\phi(x)$ as $k(x,\cdot)$, where $\phi: \mathbb{R}^d \rightarrow H_u$ maps any point in the primal space $\mathbb R^d$ to the RKHS with kernel function $k$. 
	For any two functions $g,h \in H_u$, define the inner product $\langle g,h \rangle$ as $g^Th$ and the RKHS norm as $\Vert g \Vert_{k_u}$ as $\sqrt{g^Tg}$. 
	Then, the reproducing property of the RKHS implies $f_u(x)=\langle f_u,k(x,\cdot) \rangle_{k} = \langle f_u^T,\phi(x) \rangle_{k}=f_u^T\phi(x)$ and $k(x,x^\prime)=\langle \phi(x),\phi(x^{\prime}) \rangle_{k}=\phi(x)^T \phi(x^\prime)$ for all $x,x^{\prime} \in D$. 
	Defining $\Phi_{N,j}=\left[\phi(X_j),\dots,\phi(X_{N-M+j}) \right]$ for $j=0,\dots,M$, we get the kernel matrix 
	\begin{equation}
	\begin{aligned}
	K=\sum_{i=0}^M\sum_{j=0}^M b_ib_j \Phi_{N,i} \Phi_{N,j}^T=\sum_{i=0}^M b_i\Phi_{N,i} \left(\sum_{i=0}^M b_i\Phi_{N,i}\right)^T.
	\end{aligned}	
	\end{equation}
	Further, it holds that $k(x)=\sum_{i=0}^M b_i \Phi_{N,i}\phi(x)$ and $\tilde Y=\sum_{i=0}^M b_i \Phi_{N,i}f_u$.  
	Analogue to \cite{multiarm} it holds that
	\begin{equation}
	\begin{aligned}
	&\left(\sum_{i=0}^M b_i \Phi_{N,i}\right)^T\left(\sum_{i=0}^M b_i \Phi_{N,i}\left(\sum_{i=0}^M b_i \Phi_{N,i}\right)^T+\lambda I\right)^{-1} \\ =&\left(\sum_{i=0}^M b_i \Phi_{N,i}\left(\sum_{i=0}^M b_i \Phi_{N,i}\right)^T+\lambda I\right)^{-1} \left(\sum_{i=0}^M b_i \Phi_{N,i}\right)^T.
	\end{aligned}
	\end{equation}
	Also, it holds that 
	\begin{equation}
	\left(\left(\sum_{i=0}^M b_i \Phi_{N,i}\right)^T\left(\sum_{i=0}^M b_i \Phi_{N,i}\right)+\lambda I \right)\phi(x)=\left(\sum_{i=0}^M b_i \Phi_{N,i}\right)^T k(x)+\lambda \phi(x),
	\end{equation}
	thus
	\begin{equation}
	\begin{aligned}
	\phi(x)^T\phi(x)&=k(x)^T \left(\sum_{i=0}^M b_i \Phi_{N,i}\left(\sum_{i=0}^M b_i \Phi_{N,i}\right)^T+\lambda I \right)^{-1}k(x) \\
	+&\lambda \phi(x)^T \left(\sum_{i=0}^M b_i \Phi_{N,i}\left(\sum_{i=0}^M b_i \Phi_{N,i}\right)^T+ \lambda I \right)^{-1}\phi(x),
	\end{aligned}
	\end{equation}
	which gives 
	\begin{equation}
	\begin{aligned}
	&\lambda \phi(x)^T\left(\sum_{i=0}^M b_i \Phi_{N,i}\left(\sum_{i=0}^M b_i \Phi_{N,i}\right)^T+\lambda I\right)^{-1}\phi(x) \\
	=&k(x,x)-k(x)^T(K+\lambda I)^{-1}k(x)=\sigma^2(x). 
	\end{aligned}
	\end{equation}
	Now, it holds that 
	\begin{equation}
	\begin{aligned}
	&|f_u(x)-k(x)^T(K+\lambda I)^{-1} \tilde Y| \\
	\leq & \Bigl| \phi(x)^Tf_u-k(x)^T(K+\lambda I)^{-1} \sum_{i=0}^M b_i \Phi_{N,i}f_u \Bigr| \\
	=&\Bigl| \phi(x)^Tf_u-\phi(x)^T(K+\lambda I)^{-1}\left(\sum_{i=0}^M b_i \Phi_{N,i}\right)^T \left(\sum_{i=0}^M b_i \Phi_{N,i}\right)f_u \Bigr| \\
	\leq & \Vert \lambda (K+\lambda I)^{-1} \phi(x)\Vert_{k}  \Vert f_u\Vert_{k} \\
	=&C_u \sqrt{\lambda \phi(x)^T(K+\lambda I)^{-1}\lambda I(K+\lambda I)^{-1} \phi(x)} \\
	\leq & C \sqrt{\lambda \phi(x)^T(K+\lambda I)^{-1} \phi(x)} \\
	=&C \sigma(x).
	\end{aligned}
	\end{equation}
	For the error $r$ it holds that 
	\begin{equation}
	\begin{aligned}
	&|k(x)^T(K+\lambda I)^{-1}r| \\
	=&\Bigl|\phi(x)^T \left(\sum_{i=0}^M b_i \Phi_{N,i}\left(\sum_{i=0}^M b_i \Phi_{N,i}\right)^T+\lambda I\right)^{-1}\left(\sum_{i=0}^M b_i \Phi_{N,i}\right)r\Bigr| \\
	\leq & \Big\Vert \left(\sum_{i=0}^M b_i \Phi_{N,i}\left(\sum_{i=0}^M b_i \Phi_{N,i}\right)^T+\lambda I \right)^{-\frac{1}{2}} \phi(x) \Big\Vert_{k} \\
	& \Big\Vert \left(\sum_{i=0}^M b_i \Phi_{N,i}\left(\sum_{i=0}^M b_i \Phi_{N,i}\right)^T+\lambda I \right)^{-\frac{1}{2}} \left(\sum_{i=0}^M b_i \Phi_{N,i}\right)^T r \Big\Vert_{k}  \\
	=& \sqrt{\phi(x)^T \left(\sum_{i=0}^M b_i \Phi_{N,i}\left(\sum_{i=0}^M b_i \Phi_{N,i}\right)^T+\lambda I \right)^{-1} \phi(x) } \\
	&\sqrt{\left(\sum_{i=0}^M b_i \Phi_{N,i}r\right)^T  \left(\sum_{i=0}^M b_i \Phi_{N,i}\left(\sum_{i=0}^M b_i \Phi_{N,i}\right)^T+\lambda I\right)^{-1}\sum_{i=0}^M b_i \Phi_{N,i}r} \\ 
	=& \lambda^{-\frac{1}{2}}\sigma(x)\sqrt{r^TK(K+\lambda I)^{-1}r}. \\
	\end{aligned}
	\end{equation}
	Analogous to \citet{multiarm}, we get the error 
	\begin{equation}
	\begin{aligned}
	&\|f_u(x)-k(x)^T(K+\lambda I)^{-1} Y\| \\
	\leq & \sigma(x)(C+(1+\tau)^{-\frac{1}{2}}\sqrt{r^T ((K+\tau I)^{-1}+I)^{-1} r}) \\
	\leq &\sigma(x) (C+(1+\tau)^{-\frac{1}{2}} \epsilon \sqrt{\Vert((K+\tau I)^{-1}+I)^{-1} \Vert}_2),
	\end{aligned}
	\end{equation}
	with $\lambda = 1+\tau$. 
	This yields the result.
\end{proof}
In order to obtain our results, we need a definition from integrator theory.

\paragraph{Consistency: }
A variable step size method is consistent of order $P$, if
\begin{equation}
\sum_{j=0}^{M} a_{jn}q(x_{n+j})=\sum_{j=0}^M b_{jn}q^{\prime}(x_{n+j})
\end{equation}
holds for all polynomials $q$ of degree $\leq P$ and for all grids $x_j$. 
In particular, the order of consistency determines the truncation index of the Taylor series up to which the true and approximated solution coincide. 

\begin{theorem}[Approximation error in the irregularly-sampled case] \label{approx}
	Consider $N$ points $X \in \mathbb{R}^{d \times N}$ with $X=(X_n)_{n=1 \dots N}$ with $X_n=x(t_n) \in \Omega$,  where $\Omega \subset \mathbb{R}^{d}$ is a compact subset and corresponding step sizes $h_n=t_{n+1}-t_n$.
	Assume that it holds component-wise that $|f_u^{P+1}(x)|_{\Omega} \leq L$ and $|f_u^{P+2}(x)| \leq L$ for $u=1,\dots,d$. 
	Then, it holds for $n \in 1,\dots,N$ and $u=1,\dots,d$ that
	\begin{equation}
	\Bigl|\sum_{j=0}^M a_{jn} X_{n+j} - \sum_{j=0}^M b_{jn} f_u(X_{n+j})\Bigr| \leq L M^{P+1} \frac{\max(h)^{P+1}}{(P+1)!} \sum_{j=0}^M (|a_{jn}|+|b_{jn}|)
	\end{equation}
\end{theorem}
for the local error of the multistep method with varying step sizes.
\begin{proof}	
	Developing a Taylor series around $X_n$ yields
	\begin{equation}
	\begin{aligned}
	& \Bigl|\sum_{j=0}^M a_{jn} X_{n+j} - \sum_{j=0}^M b_{jn} f_u(X_{n+j})\Bigr| \\
	=& \Bigl|\sum_{j=0}^M a_{jn} \sum_{i=0}^{\infty} \frac{\tilde h_{jn}^i}{i!}f_u^i(X_{n})- \sum_{j=0}^M b_{jn} \sum_{i=0}^{\infty} \frac{\tilde h_{jn}^i}{i!}f_u^{i+1}(X_{n})\Bigr|,
	\end{aligned}
	\end{equation} 	
	where $\tilde h_{0n}=0$ and $\tilde h_{jn}=\sum_{k=0}^{j-1} h_{n+k}$.
	For a method that is consistent of order $P$, the first $P$ summands vanish, thus it holds that 
	\begin{equation}
	\begin{aligned}
	&\bigl| \sum_{j=0}^M a_{jn} \sum_{i=0}^{\infty} \frac{\tilde h_{jn}^i}{i!}f_u^i(x)- \sum_{j=0}^M b_{jn} \sum_{i=0}^{\infty} \frac{\tilde h_n^i}{i!}f_u^{i+1}(x) \bigl| \\
	=&\bigl| \sum_{j=0}^M a_{jn} \sum_{i=P+1}^{\infty} \frac{\tilde h_{jn}^i}{i!}f_u^i(x)- \sum_{j=0}^M b_{jn} \sum_{i=P+1}^{\infty} \frac{\tilde h_{jn}^i}{i!}f_u^{i+1}(x) \bigl| \\
	\leq &\frac{L}{(P+1)!} \sum_{j=0}^M |a_{jn}|h_{jn}^{P+1}+\sum_{j=0}^M \tilde h_{jn}^{P+1}|b_{jn}| \\
	\leq &L M^{P+1}\frac{\max(h)^{P+1}}{(P+1)!} \sum_{j=0}^M (|a_{jn}|+|b_{jn}|).
	\end{aligned}
	\end{equation}
\end{proof}
Note that by assuming $f$ to be analytic, $f^P(x)$ can be estimated via the dynamics $f$. 
\begin{theorem}[Error for irregularly-sampled case]
	Consider $N$ inputs $X \in \mathbb{R}^{d \times N}$ with $X=(X_n)_{n=1 \dots N}$ and $X_n=x(t_n) \in \Omega$,  where $\Omega \subset \mathbb{R}^{d}$ is a compact subset and corresponding step sizes $h_n=t_{n+1}-t_n$. Further, consider observations $Y=(Y_n)_{n=1 \dots N}$ and $Y_n=\sum_{j=0}^M a_{jn}X_{n+j}$.
	Assume that the setting of Theorem \ref{multistepGP} and Theorem \ref{approx} holds.  
	Further, $u \in \{1,\dots,d\}$ and $f_u \in H_u$ with kernel $k$ and let $\Vert f \Vert_{k} \leq C$.
	Again, consider $\mu(x)=k(x)^T(K+\lambda I)^{-1}Y$. 
	Then, it holds that 
	\begin{equation}
	\begin{aligned}
	\Vert f_u(x)- \mu(x) \Vert & \leq \sigma(x)(C+C_{\epsilon}(1+\tau)^{-\frac{1}{2}}\sqrt{\Vert((K+\tau I)^{-1}+I)^{-1}\Vert}_2),
	\end{aligned}
	\end{equation}
	with 
	\begin{equation}
	C_{\epsilon}=LM^{P+1}\frac{\max(h)^{P+1}}{(P+1)!}(N-M)\sum_{j=0}^M(|a_{jn}|+|b_{jn}|).
	\end{equation}
\end{theorem}
\begin{proof}
	Theorem \ref{approx} yields
	\begin{equation}
	\Vert \mu(x)-f(x) \Vert \leq \sigma(x)(C+(1+\tau)^{-\frac{1}{2}}  \epsilon   \sqrt{\Vert ((K+\tau I)^{-1}+I)^{-1}}\Vert_2).
	\end{equation}
	From Theorem \ref{approx} we obtain $\epsilon=LM^{P+1}(N-M)\frac{\max(h)^{P+1}}{(P+1)!}{\sum_{j=0}^M (|a_{jn}|+|b_{jn}|)}$. 
	This yields the statement. 
\end{proof}
\paragraph{Interpretation}
In general, our error for the multistep methods is composed of two components. 
The first one corresponds to the approximation error of the GP for multistep kernels and is an extension of the standard GP error bounds \citep{multiarm}.
The second determines the error bound of the integrator. 
Intuitively, the first $P$ summands of the corresponding Taylor series are cut and the remaining error is estimated from prior knowledge about the underlying function and the multistep components. 
The results are obtained for the true RKHS of the continuous-time dynamics function.
Thus, it is assumed that the kernel hyperparameters are correct, which is not the case in practice. 
However, the results could be extended to kernel mismatch by leveraging the results in  \citet{Fiedler_Scherer_Trimpe_2021}. 

\subsection{Taylor case}
Next, we first present the results for the Taylor setting.
The strategy is as follows:
We assume that the continuous-time dynamics function is part of an RKHS.
This allows to compute the kernels for the higher-order Lie derivatives and thus, also the RKHS for a truncated Taylor expansion.
Error bounds are obtained by combining the error bound of the truncated Taylor expansion with the GP error bound. 
As before, we will derive each error bound dimension-wise. 

\begin{theorem}[Approximation error]\label{error}
	Consider a Taylor approximation up to order $P$ given by $x_u^P(x,h)=\sum_{k=0}^P \frac{h^k}{k!}f_u^k(x)$. 
	Assume that $x(s) \in \Omega$ for $s \in [t,t+h]$ with $\Omega \subset \mathbb{R}^d$ a compact subset. 
	Further, assume that it holds componentwise $|f_u^{P+1}(x)|_{\Omega} \leq E$, where $u=1,\dots d$.
	Then for the approximation error $x_u(t+h)-x_u^P(x,h)$ it holds with the midpoint rule that
	\begin{equation}
	|x(t+h)-x_u^P(x,h)|\leq \epsilon,  
	\end{equation}	
	where $\epsilon= \frac{h^{P+1}}{(P+1)!}E$.
\end{theorem} 

\begin{proof}
	It holds that 
	\begin{equation}
	|x_u(t+h)-x_u^P(x,h)|=|\sum_{k=P+1}^{\infty}\frac{h^k}{k!}f_u^k(x)| \leq \frac{h^{P+1}}{(P+1)!}E.
	\end{equation}
	Thus, the result follows. 
\end{proof} 
Note, that for an analytic bounded $f$, the bound for the Lie-derivatives can be estimated.
Next, we construct an RKHS that contains the function of interest. 
Again, each dimension $u=1,\dots,d$ is addressed via an independent Gaussian process with kernel $k$.
\begin{theorem}[Taylor approximation]\label{taylor}
	Consider $u \in \{1,\dots,d\}$ and let $f_u^l \in H^l_u$ for  $l=1,\dots,P$
	with kernel $k_l(x,y)$ and $\Vert f_u^l \Vert_{k_l} \leq C_l$. 
	Consider $N$ training data $x(t_n) \in \Omega$ for $n=1,\dots,N$ where $\Omega \subset \mathbb R^d$ is a compact set. This yields $N-1$ inputs $X=(X_n)_{n=1,\dots,N} \in \mathbb{R}^{d \times N-1}$ with $X_n=x(t_n)$ and observations $Y \in \mathbb{R}^{N-1}$ with $Y=(Y_n)_{n=1 \dots N-1} \in \Omega$ and $Y_n=x_u(t_{n+1})-x_u(t_n)$ and corresponding step sizes $h_n$. 
	Consider the mean approximation for $f_u=f_u^1(x)$ with $\mu(x)=k(x)(K+\lambda I)^{-1}Y$. 
	With the assumptions and definitions from Theorem \ref{error} it holds that 
	\begin{equation}
	|f_u(x)-\mu(x)| \leq \sigma(x)\left(C+C_{\epsilon}(1+\tau)^{-\frac{1}{2}}\sqrt{\Vert((K+\tau I)^{-1}+I)^{-1}\Vert_2}\right), 
	\end{equation}
	where
	\begin{equation}
	C_{\epsilon}=(N-1) \epsilon,
	\end{equation}
	$C=\sqrt{\sum_{l=0}^P (C_l)^2}$, $\epsilon=\frac{\text{max}(h)^{P+1}}{(P+1)!}E$ and $\lambda=1+\tau$. 
\end{theorem} 
\begin{proof}
	Consider the extended RKHS $H_u:\mathbb{R}^P \times \mathbb{R}^d \rightarrow \mathbb{R}$ with kernel function $\tilde k(x,y,h,i)=\sum_{l=1}^P h_l i_l k^l(x,y)$.
	Further, consider the RKHS function 
	\begin{equation}
	F_u^P(x,h)=\sum_{l=1}^P h_l f_u^l(x).
	\end{equation}
	Define $\tilde h=[1,0,\dots,0]$. 
	Conditioning $F_u^P$ on the inputs $X$ and corresponding stepsizes $[h_n,\dots,h_n^P]$, we obtain $\mu$ and $\sigma$ as defined in the method section of the main paper. 
	Similar to Theorem 2 from \citet{multiarm} and the proof of Thm. \ref{approx}, we obtain
	\begin{equation}
	\begin{aligned}
	& \Vert F_u^p(\tilde h, x)-\mu(x) \Vert \\
	\leq & \sigma(x)\left(\Vert F_u^P \Vert_{\tilde k}+(1+\tau)^{-\frac{1}{2}}(N-1) \epsilon\sqrt{\Vert ((K+\tau I)^{-1}+I)^{-1} \Vert_2} \right).
	\end{aligned}
	\end{equation}
	It holds that $\Vert F_u^P \Vert_{k}=\sqrt{\sum_{l=0}^P (C_l)^2}$ since the linear component has norm 1. 
	Further, it holds that $\epsilon=\frac{h^{P+1}}{(P+1)!}E$ with Theorem \ref{error}.
	This yields the result  
\end{proof}

\emph{Note:}
Assuming that $f_u^1 \in H_u^1$ for $u=1\dots d$ with smooth enough kernel, we can derive the higher-order Lie-derivatives by applying Eq. \eqref{eq:Lie_kernel}.

\paragraph{Interpretation}
Similar as for the multistep method, our error is composed of two components.
The first one corresponds to the approximation error of the GP with Taylor kernels and is an extension of the standard GP error bounds \citep{multiarm}.
The second one determines the error bound of the integrator. 
Intuitively, the first $P$ summands of the corresponding Taylor series are cut and the remaining error has to be estimated from prior knowledge about the function.
Again, we assume that the kernels for each GP are chosen correctly.
However, as before the results can be extended to kernel mismatch by leveraging the results in \citet{Fiedler_Scherer_Trimpe_2021}.

%% file: add_results.tex
\section{Additional results}
In this section, we present additional results. 
In particular, we add results for integration with the same integrator used during training for completeness. 
Further, we add results for the mean predictions, demonstrating that we achieve similar or even higher accuracy than with decoupled sampling.
For the Van-der-Pol oscillator, we also add results with a different time irregularity.

\subsection{Mean predictions} 
In the main paper, we present results for the decoupled sampling predictions with all integrators. 
Here, we provide additional results for mean predictions. 
Table \ref{DHO} shows the results for the DHO system, Table \ref{VDP} for der Van-der-Pol oscillator and \ref{real_mass} for the real mass-spring system.
The results are consistent with the results for DS predictions. 
With increasing integrator order, we obtain better ODE approximations.
In general, the accuracy of the mean predictions is slightly higher with respect to the DS predictions. 

\begin{table}
	\centering
	\begin{tabular}{rcc}
		\hline
		Train Integrator & order & MSE \\ \hline
		AB & 1  & 0.932 (0.034) \\  
		AB & 2 &2.637215(0.009) \\  
		AB & 3 & 0.037 (0.008) \\   
		AM/BDF & 1 &1.168 (0.038) \\  
		AM & 2 &0.015 ($9\cdot 10^{-4}$) \\ 
		AM & 3 & 0.021 (0.001) \\  
		BDF & 2 & 0.006 ($7 \cdot 10^{-4}$) \\ 
		BDF & 3 &  0.003 ($2 \cdot 10^{-4}$) \\ 
	\end{tabular}
	\caption{Damped harmonic oscillator: Mean predictions with RK4(5) rollouts.}
	\label{DHO}
\end{table}
\begin{table}
	\centering
	\begin{tabular}{rcc}
		\hline
		Train Integrator & order  & MSE \\ \hline
		AB & 1  & 2.19 (0.008) \\ 
		AB & 2  & 0.018 (0.0) \\ 
		AB & 3  & 0.02 (0.00) \\ 
		AM/BDF & 1  & 19.815 (0.069) \\ 
		AM  & 2 & 0.011 (0.001) \\ 
		AM  & 3 & 0.0167 ($2 \cdot 10^{-6}$) \\ 
		BDF & 2 & 0.039 (0.002) \\
		BDF & 3 & 0.027 (0.007) \\ 
		Taylor & 2 & 0.005 (0.001) \\ 
		Taylor & 3 & 0.005 (0.001) \\ 
	\end{tabular}
	\caption{Van-der-Pol oscillator: Mean predictions with 50\% irregularity and RK4(5) rollouts.}
	\label{VDP}
\end{table}

\begin{table}[h!]
	\centering
	\begin{tabular}{rcc}
		\hline
		Train Integrator & order & MSE \\ \hline
		AB & 1 & 0.510 ($4 \cdot 10^{-7}$) \\ 
		AB & 2 & 0.009 ($7 \cdot 10^{-6}$) \\ 
		AB & 3 & 0.002 ($10^{-6}$) \\ 
		AM/BDF & 1 & 662.73 (1.61) \\ 
		AM & 2 & 0.003 ($3 \cdot 10^{-6}$) \\ 
		AM & 3 & 0.002 ($2 \cdot 10^{-6}$) \\ 
		BDF & 2 & 0.019 ($\cdot 10^{-6}$) \\ 
		BDF &3 & 0.012 (0.0) \\ 
	\end{tabular}
	\caption{Real mass-spring system: Mean predictions with 50\% irregularity and RK4(5) rollouts.}
	\label{real_mass}
\end{table}

\subsection{Predictions with training integrator}
In the main paper, we presented predictions with the adaptive step size integrator RK4(5).
Here, we present additional results for predictions with the training integrator. 
Thus, we train and predict on the identical discrete-time system and do not get discretization errors during predictions.
The results on regularly-sampled grids as Table \ref{DS:DHO} and Table \ref{DS:RM} demonstrate that the integrators provide similar accuracy when being rolled out with the identical integrator. 
Thus, the differences in performance are indeed caused by the fact that higher-order integrators provide more accurate ODE models. 
The results for the Van-der-Pol oscillator in Table \ref{DS:VDP} on an irregular grid instead demonstrate the original trend again.
Higher order integrators provide better results here.  
This demonstrates that on irregularly sampled grids, a lower order discretization is not sufficient. 
Intuitively, lower-order approximations as the explicit Euler method can only explain a predefined discretization level but no variation in the step sizes.

\begin{table}[h!]
	\centering
	\begin{tabular}{rcc}
		\hline
		Integrator & order & MSE  \\ \hline
		AB & 1 & 0.287 (0.082) \\ 
		AB & 2 & 0.034 (0.015) \\ 
		AB & 3 & 0.052 (0.021 )  \\ 
		AM/BDF & 1 & 0.014 (0.004 ) \\ 
		AM & 2 & 0.023 (0.014 )\\ 
		AM & 3 & 0.036 (0.023 )\\ 
		BDF & 2 & 0.018 (0.024 ) \\ 
		BDF & 3 & 0.006 ($8 \cdot 10^{-4}$) \\ 
	\end{tabular}
	\caption{Damped harmonic oscillator: DS predictions with training integrator.}
	\label{DS:DHO}
\end{table}
\begin{table}[!ht]
	\centering
	\begin{tabular}{rcc}
		\hline
		Integrator & order & MSE \\ \hline
		AB & 1  & 0.244(0.099) \\  
		AB & 2  & 0.059(0.025) \\  
		AB & 3  & 0.142(0.085) \\  
		AM/BDF & 1  & 0.398(0.123)\\  
		AM & 2  & 0.002(0.003)  \\  
		AM & 3  & 0.049(0.070) \\  
		BDF & 2 & 0.088(0.055) \\  
		BDF & 3 & 0.013(0.010) \\  
		Taylor & 2 & $0.001$ ($5 \cdot 10^{-4}$) \\ 
		Taylor & 3 & $0.003$ ($0.003$) \\ 		
	\end{tabular}
	\caption{Van-der-Pol oscillator: DS predictions with 50\% irregularity and training integrator.}
	\label{DS:VDP}
\end{table}

\begin{table}[!ht]
	\centering
	\begin{tabular}{rcc}
		\hline
		Integrator & order  & MSE  \\ \hline
		AB & 1 & 0.031(0.041)  \\  
		AB & 2 & 0.016(0.018) \\  
		AB & 3 & 0.005(0.007)  \\  
		AM/BDF & 1 & 0.037(0.050) \\  
		AM & 2 & 0.012(0.008)  \\  
		AM & 3 & 0.017(0.017)  \\  
		BDF & 2 & 0.053(0.070) \\  
		BDF & 3 & 0.038(0.033)  \\  
	\end{tabular}
	\caption{Real spring system: DS predictions with training integrator.}
	\label{DS:RM}
\end{table}

\subsection{Additional experiments} 
Here, we present additional results. 
We present results for the Van-der-Pol oscillator with a different level of irregularity.

\paragraph{Van-der-Pol oscillator: }
\label{VDP_30} shows results for the Van-der-Pol oscillator with 30\% irregularity.
Predictions are calculated with decoupled sampling and an adaptive step-size integrator. 
The results show a similar trend as the ones with 50\% irregularity. 
For Taylor, the results with 30\% irregularity are outperformed by the ones with 50\%.
Intuitively, this is since more irregularity in the data restricts the freedom of each GP component to take over the wrong parts of the learning task.

\begin{table}[!ht]
	\centering
	\begin{tabular}{rcc}
		\hline
		Integrator &  order & MSE \\ \hline
		AB & 1 & 1.09 (0.05) \\ 
		AB & 2 & 0.052 (0.013) \\ 
		AB & 3 & 0.098 (0.052) \\ 
		AM/BDF & 1 & 25.613 (17.010) \\ 
		AM & 1 & 0.010 (0.001) \\ 
		AM & 2 & 0.008(0.002) \\ 
		BDF & 2 & 0.20 (0.262) \\ 
		BDF & 3 & 0.092 (0.071) \\ 
		Taylor & 2 & 0.08 (0.034) \\  
		Taylor & 3 & 0.05 (0.034) \\ 
	\end{tabular}
	\caption{DS predictions with RK4(5) for VDP with 30 \% irregularity.}
	\label{VDP_30}
\end{table}


\subsection{Additional information}
\begin{figure*} 
	\centering 
	\begin{subfigure}[htb]{0.49\textwidth}
		\centering
		\includegraphics[width= \textwidth]{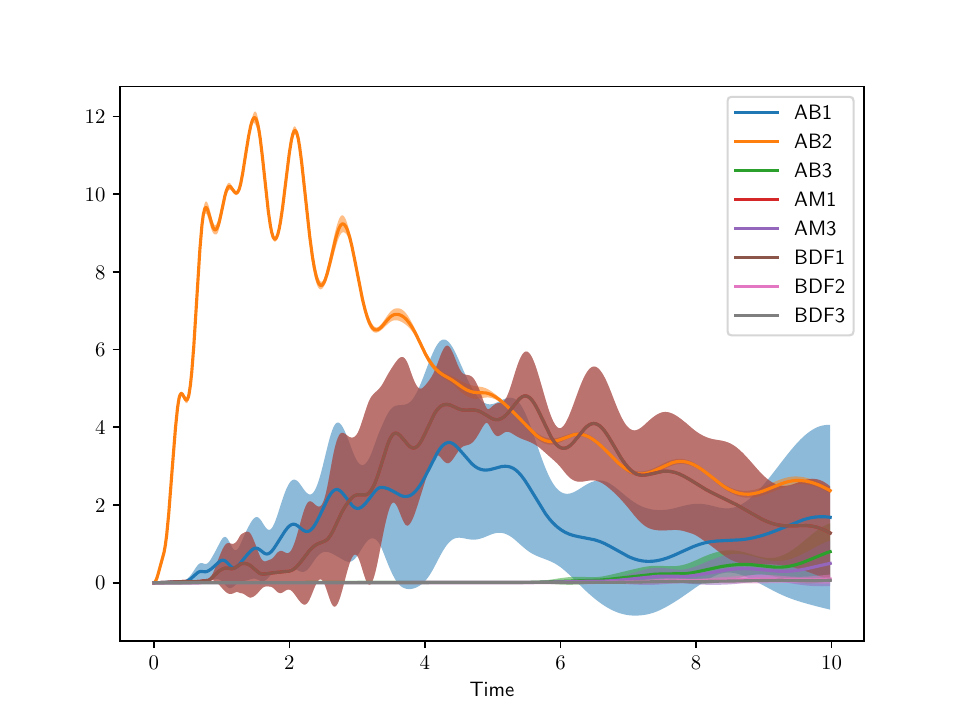}
		\caption{Damped harmonic oscillator} \label{subfig:DHO}
	\end{subfigure}
	\begin{subfigure}[htb]{0.49\textwidth}
		\centering	
		\includegraphics[width= \textwidth]{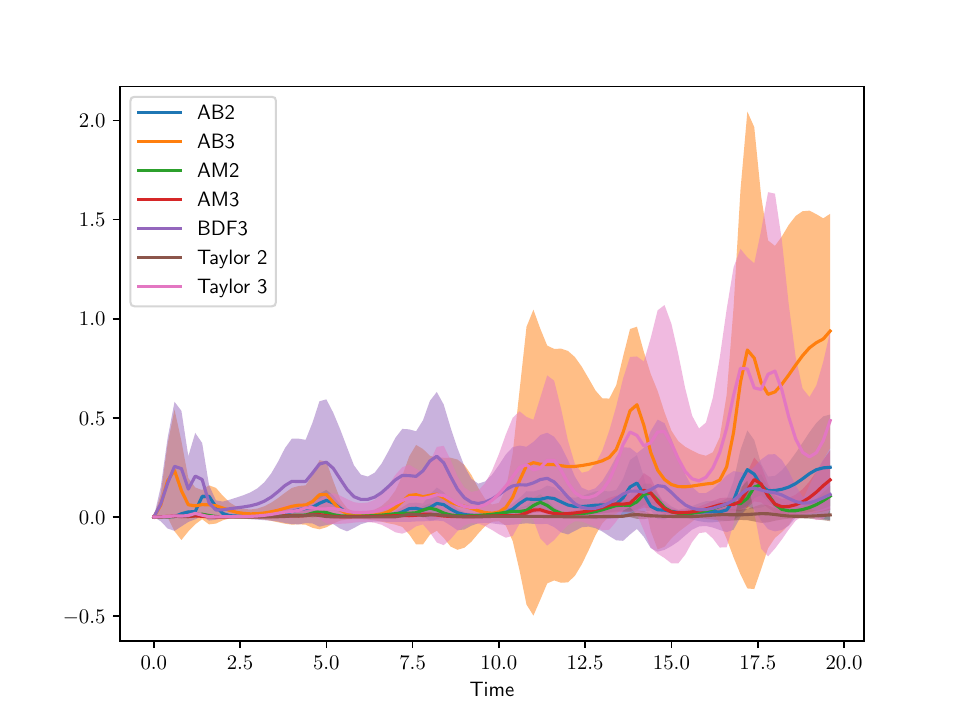}
		\caption{Van-der-Pol oscillator} \label{subfig:vdp}
	\end{subfigure}	
	\caption{RMSE over time for the DHO in Fig. \ref{subfig:DHO} and the Van-der-Pol oscillator in Fig. \ref{subfig:vdp}.}
	\label{fig:losses}
\end{figure*}

\begin{figure*} 
\centering
	\begin{subfigure}[htb]{0.49\textwidth}
	\centering
	\includegraphics[width= \textwidth]{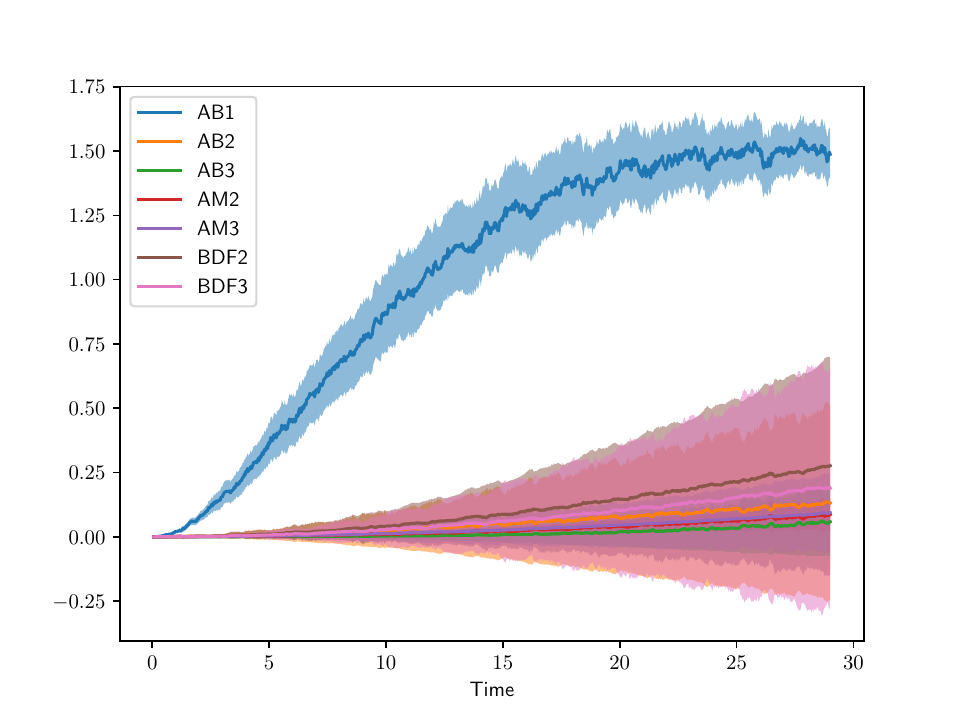}
\end{subfigure}
\caption{RMSE over time for real spring system.} 
	\label{fig:real_lin}
\end{figure*}

In this section, we provide additional plots for the RMSE over time for all experiments in the main paper.
We further provide restructured tables that emphasize the order of the underlying integrator. 
In the main paper, we provided tables omitting to note experiments twice due to space constraints. 
Here, we provide a restructured format that emphasizes the relation between integrator order and result.
Each table provides results for the MSE.
See DHO results in Table \ref{DHO}, VDP results in Table \ref{VDP} and results for the real spring system in Table \ref{real_spring}.
Further, we provide results for the error accumulation over time. 
See results for the DHO system in Fig. \ref{subfig:DHO}, for the Van-der-Pol oscillator in Fig. \ref{subfig:vdp} and for the real spring system in Fig. \ref{fig:real_lin}.

\begin{table}[!ht] 
	\centering
	\begin{tabular}{rccc}
		\hline
		Integrator/ order \vline & 1 & 2 & 3   \\ \hline
		Baseline GP-ODE \vline &  & 0.187 (0.091) &  \\   \hline
	    AB \vline &  0.750 (0.232) & 2.62 (0.011) &  0.062 (0.027)  \\ 
		AM \vline &  1.175  (0.060)  &  0.027 (0.015)  &  0.043 (0.023) \\
		BDF \vline & 1.175  (0.060) & 0.009 (0.006) & \textbf{0.006} (0.001) \\ \hline

	\end{tabular}
	\caption{DHO: DS predictions with RK4(5) (restructured).}
	\label{DHO}
\end{table}
 
\begin{table}[!ht] 
	\centering
	\begin{tabular}{rccc}
		\hline
		Integrator/ order \vline & 1 & 2 & 3   \\ \hline
		Baseline GP-ODE \vline &  & 0.043 (0.028) &  \\   \hline
		AB (Baseline) \vline &  1.843 (0.141) & 0.030 (0.007) &  0.114 (0.058) \\ 
		AM \vline &  34.578 (18.372)  & 0.015 (0.003)  &  0.016 (0.005) \\
		BDF \vline & 34.578 (18.372) & 17.753 (30.677) & 0.063 (0.037) \\ 
		Taylor \vline & 1.843 (0.141)  &  0.010 (0.004) & 0.006 (0.003)  \\ \hline 

	\end{tabular}
	\caption{VDP: DS predictions with RK4(5) (restructured).}
	\label{VDP}
\end{table} 
 
\begin{table}[!ht] 
	\centering
	\begin{tabular}{rccc}
		\hline
		Integrator/ order \vline & 1 & 2 & 3   \\ \hline
		Baseline GP-ODE \vline &  & 0.163 (0.093)  &  \\  \hline 
		AB \vline &  0.502 (0.029)  &  0.026 (0.037) &   \textbf{0.007} (0.009)   \\ 
		AM \vline & 420.5 (422)  &  0.014 (0.004)  & 0.016 (0.014) \\
		BDF \vline &  420.5 (422) & 0.050 (0.037) &  0.035 (0.041)\\ \hline

	\end{tabular}
	\caption{real spring system: DS predictions with RK4(5) (restructured).}
	\label{real_spring}
\end{table} 
 
\subsection{Additional results for Taylor}
In this section, we run experiments for the Taylor integrators.
In contrast to the results in the main paper, the GP for each Lie-derivative is trained with dependent hyperparameters and adapted kernel functions. 
The results are displayed in Table \ref{VDP_taylor_mean} for mean predictions with RK4(5) and in Table \ref{VDP_mean_train} for mean predictions with training integrator. 
\begin{table}[!ht]
	\centering
	\begin{tabular}{rccc}
		\hline
		Train Integrator & order & VDP 30 \% & VDP 50 \%  \\ \hline
		Taylor & 2 & 0.01 (0.002) &  0.002 $6 \cdot 10^{-4}$  \\ 
		Taylor & 3 & 0.007 (0.003) & 0.03 (0.004)  \\ \hline
	\end{tabular}
	\caption{VDP with adapted Taylor kernels. Mean predictions and RK4(5) rollouts.}
	\label{VDP_taylor_mean}
\end{table}

\begin{table}[!ht]
	\centering
	\begin{tabular}{rccc}
		\hline
		Train Integrator & order & VDP 30 \% & VDP 50 \%  \\ \hline
		Taylor & 2 & $3 \cdot 10^{-4} (10^{-6})$  &  $6 \cdot 10^{-4} (4 \cdot 10^{-6})$  \\ 
		Taylor & 3 & $3 \cdot 10^{-4} (10^{-5})$ & $3 \cdot 10^{-4} (6 \cdot 10^{-5})$  \\ \hline
	\end{tabular}
	\caption{VDP with adapted Taylor kernels. Mean predictions and rollouts with training integrator.}
	\label{VDP_mean_train}
\end{table}

%% file: exp_setup.tex
\section{Experimental details}
In this section, we describe the experimental setup for all experiments.
In particular, we state how the GPs are trained and add details on the computation of decoupled sampling.

\subsection{Systems}
We provide the equations for the simulated systems that were missing in the main paper. 

\paragraph{Damped harmonic oscillator (DHO): }
The damped harmonic oscillator models an oscillating system subject to a damping or friction force that slows the system exponentially, upon reaching equilibrium. 
Here, we consider a damped harmonic oscillator with cubic dynamics as considered in \citep{multisteperror}. 
\begin{equation} \label{eq:dho_system}
\begin{split}
\dot{x} &= -0.1 x^3 + 2.0 y^3 \\
\dot{y} &= -2.0 x^3 - 0.1 y^3. 
\end{split}
\end{equation}
\paragraph{Van-der-Pol oscillator (VDP): } 
We consider the Van-der-Pol oscillator \citep{cveticanin2013}.
\begin{equation} \label{eq:vdp_system}
\begin{split}
\dot{x} &= y \\
\dot{y} &= - x + 0.5 y (1 - x^2), 
\end{split}
\end{equation}
where it has been choosen $\mu = 0.5$.

\subsection{Experimental setup}
Here, we provide details on the setup of our experiments. 
The experiments were conducted on an internal cluster using CPUs.
\paragraph{Decoupled sampling: }
All GPs are trained with ARD kernels.
For the ARD kernel $k(x,x')=\sigma_f^2 \exp \left(-\sum_{i=1}^d \frac{(x_i-x_i')^2}{2 l_i^2}\right)$, it holds that 
\begin{equation} 
\phi_i(x)=\sqrt{\frac{\sigma_f^2}{S}}(\cos x^T\omega_i,\sin x^T \omega_i).
\end{equation} The parameters $\omega_i$ are sampled proportional to the spectral density of the kernel $\omega_i \sim \mathcal{N}(0,\Lambda^{-1})$, where $\Lambda = \textrm{diag}(l_1^2,l_2^2,\dots,l_d^2)$. For the ARD kernel this results in $2S$ feature maps $\phi(x) \in \mathbb{R}^{2S}$. Weights $w \in \mathbb{R}^{2S}$ are sampled from a standard normal $w_i \sim \mathcal{N}(0,1)$ (see \citep{hegde2021bayesian}).

For all experiments, we use 256 random Fourier features. 
For predictions, we sample trajectories from the posterior and compute their statistical mean and variance. 
For the damped harmonic oscillator, we sample 256 trajectories, for the Van-der-Pol oscillator and the real linear system, we sample 256 trajectories. 

\paragraph{Hyperparameter initialization: }
We initialize all hyperparameter with the explicit Euler hyperparameters.
To this end, we first train a GP with the GPy library \citep{gpy2014} on the Euler method. 
The lengthscale is then initialized with the pretrained lengthscale. 

\paragraph{Activation function: }
For the GP parameters and the observation noise, we use $\exp$ activation functions to obtain positive values.